\documentclass[12pt]{elsarticle}
\usepackage{amssymb}
\usepackage{amsmath}
\usepackage{booktabs}
\usepackage{multirow}
\usepackage{subcaption}
\usepackage[table,dvipsnames]{xcolor}
\usepackage{subcaption}
\usepackage{caption}
\usepackage{rotating}
\usepackage{needspace}
\usepackage[ruled,vlined,linesnumbered]{algorithm2e}
\usepackage{dsfont}
\usepackage[displaymath,right]{lineno}
\usepackage{hyperref}
\usepackage{cleveref}

\journal{}

\newcommand{\eacc}{\text{err}_m^\text{\normalfont(ACC)}}
\newcommand{\erps}{\text{err}_m^\text{\normalfont(aRPS)}}
\newcommand{\alphacc}{\alpha_m^\text{\normalfont(ACC)}}
\newcommand{\alphrps}{\alpha_m^\text{\normalfont(aRPS)}}

\newcounter{myalg}
\begin{document}

\begin{frontmatter}

\title{\texttt{ADABORD}: a novel AdaBoost approach for ordinal classification}

\author[imibic,pduco]{Rafael Ayllón-Gavilán}
\author[uloyola]{Francisco José Martínez-Estudillo}
\author[uco]{David Guijo-Rubio\corref{cor}}\cortext[cor]{Departmento de Ciencia de la Computación e Inteligencia Artificial, Universidad de Córdoba, Campus de Rabanales, Ctra. N-IVa, Km. 396, Córdoba, 14071.} \ead{dguijo@uco.es}
\author[uco]{César Hervás-Martínez}
\author[uco]{Pedro A. Gutiérrez}

\affiliation[imibic]{organization={Department of Clinical-Epidemiological Research in Primary Care, IMIBIC},
            addressline={Avda. Menéndez Pidal S/N},
            city={Córdoba},
            postcode={14004},
            country={Spain}}

\affiliation[pduco]{organization={Programa de doctorado en Computación Avanzada, Energía y Plasmas, Universidad de Córdoba},
            addressline={Campus de Rabanales, Ctra. N-IVa, Km. 396},
            city={Córdoba},
            postcode={14071},
            country={Spain}}

\affiliation[uloyola]{organization={Department of Quantitative Methods, Universidad Loyola Andalucía},
            addressline={C. Escritor Castilla Aguayo, 4},
            city={Córdoba},
            postcode={14004},
            country={Spain}}

\affiliation[uco]{organization={Departamento de Ciencia de la Computación e Inteligencia Artificial, Universidad de Córdoba},
            addressline={Campus de Rabanales, Ctra. N-IVa, Km. 396},
            city={Córdoba},
            postcode={14071},
            country={Spain}}
            
\begin{abstract}
Ordinal Classification (OC) deals with classification tasks where the classes follow a natural order. Despite the progress in OC, many existing approaches fail to fully leverage the ordinal information, treating the problem as nominal classification and thereby losing performance potential. In this work, \texttt{ADABORD}, an AdaBoost framework specifically designed for ordinal classification problems, is introduced. The ordinal nature of the classes is incorporated into two key components of the well-known AdaBoost algorithm: 1) the base estimator, where decision trees with the ordinal Gini splitting criterion are proposed; 2) the error function used to update sample weights at each stage and the weights of the classifier in the final ensemble model, given by the absolute ranked probability score, a measure that accounts for both the ordering and the distance between classes. \texttt{ADABORD} is extensively compared against seven state-of-the-art methods on the TOC-UCO repository, the largest benchmark collection for OC to date. The experimental results, supported by statistical analysis, show that \texttt{ADABORD} significantly outperforms competing methods, particularly on datasets with five or more classes, where the ordinal structure becomes more pronounced. Source code, along with all experimental protocols, is publicly available to ensure reproducibility and facilitate future research in OC.
\end{abstract}

\end{frontmatter}

\section{Introduction} \label{sec:intro}

Ordinal Classification (OC) \cite{agresti2010analysis}, also referred to as \textit{ordinal regression}, is a type of classification problem in which the output categories exhibit an inherent order relationship among them. Conceptually, OC lies between standard classification and regression. Unlike regression, where the target variable shows explicit values, OC deals with discrete categories whose pairwise distances are typically unknown and probably non-uniform \cite{gutierrez2016OrdinalRegressionMethods}. In contrast to nominal (standard) classification, where all misclassification errors are treated with equal severity, OC considers a relative severity of errors based on the ordinal structure of the target variable. That is, the cost of a misclassification increases with the ordinal distance between the predicted and actual classes. Although traditional nominal classification methods are often used as baselines in OC scenarios, they fail to exploit the order of the labels. In contrast, ordinal-specific techniques can leverage this structured information to achieve improved predictive performance \cite{delgado2025ordmap}. Moreover, ordinal problems also exhibit a distinctive characteristic, which is a high inherent degree of class imbalance, as instances belonging to the extreme classes are naturally more difficult to obtain. There are several approaches for OC \cite{gutierrez2016OrdinalRegressionMethods}, including simplifying the problem to related tasks (nominal classification or regression), ordinal binary decompositions or threshold models. However, as discussed later, OC has received less attention in ensemble learning research.

Among ensemble classification techniques, Adaptive Boosting (AdaBoost) has played a central role due to its strong empirical performance and solid theoretical foundations. From a statistical learning perspective, AdaBoost can be interpreted as a forward stage-wise additive modelling procedure that minimises an exponential loss function in the function space \cite{freund1997decision}. Under this view, the ensemble is constructed sequentially by adding weak learners that approximate the negative functional gradient of the loss with respect to the current model, resulting in a form of functional gradient descent. The multiclass SAMME extension \cite{zhu2009multi} preserves this interpretation by optimising a multiclass exponential loss and producing a weighted additive classifier whose learning dynamics are fully determined by the choice of the loss function.

Despite its success, the original AdaBoost algorithm, and its multiclass variants, are not explicitly designed for OC problems. The exponential loss underlying AdaBoost treats all misclassification errors uniformly, regardless of their relative distance in the ordered label space. Consequently, the functional gradient descent process does not exploit ordinal relationships and may converge to solutions that are optimal in terms of nominal accuracy but suboptimal when evaluated using ordinal-aware performance criteria. This mismatch between the optimisation objective and the problem structure becomes increasingly pronounced as the number of ordinal categories grows, where the severity of misclassification errors inherently depends on an ordinal distance \cite{gutierrez2016OrdinalRegressionMethods}.

A further critical issue is the design of score-free algorithms, that is, methods that do not rely on assigning arbitrary numerical values or assuming equal spacing between ordinal categories. Many OC approaches implicitly encode labels as integers and optimise regression-based or distance-dependent losses, thereby introducing metric assumptions that are often unjustified in real-world applications. In the context of boosting, such assumptions directly shape the loss landscape and the resulting gradient directions, potentially biasing the learning process. From a theoretical point of view, score-free OC methods are therefore preferable, as they preserve label ordering without imposing artificial metric structure. Representations based on cumulative probabilities constitute an adequate alternative, allowing ordinal information to be incorporated while remaining agnostic to inter-class distances \cite{epstein1969scoring}.

Within this context, ordinal ensemble learning remains a relatively underexplored area, particularly in comparison with the extensive literature on nominal boosting methods. While several adaptations of AdaBoost to ordinal settings have been proposed, such as OCEAn \cite{vega2021ocean} or OEAdaBoost \cite{um2023adaptive}, most modify either the loss function or the base learner in isolation and often rely on implicit distance assumptions between classes.

In this work, \texttt{ADABORD}, an AdaBoost framework specifically designed for OC, is proposed, integrating ordinal information in both the base classifier and the error function. Specifically, in this proposal an OC-adapted version of the Decision Tree Classifier (DTC) that uses the Ordinal Gini (OGini) splitting criterion \cite{piccarreta2008classification} is considered. On the other hand, regarding the error function, a modified version of the Ranked Probability Score (RPS) \cite{epstein1969scoring} is presented, known as absolute RPS (aRPS). The aRPS metric computes the distance between the actual and the predicted cumulative distributions of the target variable. It is important to note that both RPS and aRPS are strictly proper scoring rules for ordered categorical outcomes, ensuring optimality in expectation when the predicted cumulative probabilities are consistent with the underlying data-generating distribution \cite{gneiting2007strictly}. Moreover, it can be regarded as score-free in the sense that it does not rely on assigning arbitrary numerical values to the categories nor on assuming equal spacing between ordinal levels.

In this way, by employing the DTC learner coupled with the OGini splitting criterion and the aRPS error measure based on cumulative probability distributions, the boosting optimisation process in \texttt{ADABORD} is aligned with the intrinsic structure of ordinal targets, while remaining fully score-free.

The main contributions of this work can be summarised as follows:
\begin{itemize}
    \item The integration of ordinal decision trees as base learners within a boosting framework, where the splitting process explicitly accounts for the ordered structure of class labels through the OGini impurity criterion. By embedding ordinal-aware decision trees into a sequential boosting scheme, ordinal information is exploited at both the feature selection and ensemble construction levels.
    \item The adoption of the aRPS as the error function within the boosting framework, replacing the standard misclassification-based loss. The aRPS measures discrepancies between predicted and observed cumulative distributions, thereby respecting the ordinal nature of the target variable without assuming explicit distances between categories.
    \item The combination of ordinal decision trees with the aRPS error function in the \texttt{ADABORD} approach leads to strong empirical performance on ordinal classification problems with five or more classes, which commonly exhibit class imbalance, particularly at the extremes of the ordinal scale.
\end{itemize}

In addition to the core methodological advances described above, this work includes several practical outcomes aimed at ensuring a rigorous and reproducible empirical evaluation. The proposed method, \texttt{ADABORD}, is compared against seven competitive state-of-the-art techniques, including the nominal version of AdaBoost, a competitive AdaBoost-based ordinal classifier (OEAdaBoost) \cite{um2023adaptive}, and three representative nominal approaches drawn from different methodological families. For this, a comprehensive experimental study is conducted using the TOC-UCO repository \cite{tocuco}, which comprises $46$ ordinal datasets spanning diverse application domains and exhibiting heterogeneous characteristics. Parametric statistical tests are performed to assess whether the differences observed between the model performances are statistically significant. Finally, to support reproducibility and facilitate future research, the full implementation of the proposed method, together with detailed instructions for reproducing the experiments and the results obtained by all compared methodologies, is publicly available\footnote{\url{https://github.com/ayrna/adabord}}.

The remainder of this paper is organised as follows. \Cref{sec:related} analyses previous advances in the field of ordinal classification, focusing on ensemble methods. \Cref{sec:methodology} defines the problem addressed in this work and presents the proposed approach, named \texttt{ADABORD}. \Cref{sec:expsetup} describes the experimental setup, including the benchmarking datasets, data partitions, baseline methods used for comparison, and the performance metrics employed. \Cref{sec:results} reports the experimental results of the proposed \texttt{ADABORD} approach, providing a comprehensive comparison against seven existing methodologies. In addition, the results are statistically analysed in \Cref{subsec:statistical}. Finally, this work concludes with some remarks (\Cref{sec:conclusions}), followed by a discussion of the limitations and potential directions for future research (\Cref{sec:limitations}).

\section{Related works} \label{sec:related}

Over the past decade, the OC field has received increasing attention in the research community \cite{gutierrez2016OrdinalRegressionMethods}. This growing interest is driven not only by the wide range of application domains in which OC techniques have proven useful, but also by the increasing need for specialised learning methodologies. In medicine, for instance, ordinal classifiers have been applied to determine the degree of affectation in Parkinson's disease from 3D images \cite{duran2021ordinal}, as well as to grade diabetic retinopathy and prostate cancer from diagnostic images \cite{TOLEDOCORTES2022105472}. In the meteorological domain, OC approaches have been employed to predict low-visibility events caused by fog \cite{guijo2018prediction} and to estimate wind speed in wind farms \cite{pelaez2024general}. The financial sector has also benefited from OC techniques, which have been used to support strategic planning in financial companies \cite{solares2022multicriteria} and to estimate corporate credit rating levels \cite{GOLDMANN20241111}. In industrial applications, OC has been used to determine the relative importance of users' perspectives on electric vehicle purchases \cite{kucuksari2023new}, as well as to estimate energy flux from ocean buoy measurements \cite{gomez2024orfeo}. Finally, ordinally structured potential categories (e.g. ``weak'', ``moderate'', ``good'', and ``very good'') can be used to characterise geothermal resources \cite{mirfallah2025geothermal}, which is essential for strategic and sustainable energy development. Nevertheless, the increasing diversity and complexity of these applications demand the development of novel OC methodologies, such as those discussed in the following paragraphs and the one presented in this work, \texttt{ADABORD}.

Among the wide range of OC methodologies published in the literature, two families of methods stand out as having made the most significant contributions to the development of ordinal classification: neural network-based approaches and decision tree-based approaches. In the context of neural networks, Lázaro et al. \cite{lazaro2023neural} introduced a Bayesian training formulation, where a loss function based on an estimate of the Bayesian classification cost is proposed. This formulation employs a Parzen window estimator fitted to a thresholded decision scheme, allowing flexible specification of the relative importance of classification errors across different classes. Additionally, Okuno et al. \cite{okuno2024interpretable} presented an interpretable neural network-based model built upon the non-proportional odds framework. Their approach achieves a balance between predictive flexibility and model interpretability in OC. 

Concerning tree-based methodologies, Marudi et al. \cite{marudi2024decision} proposed a decision tree framework that generalises the traditional entropy-based splitting criterion, making it applicable across various tree-based OC models. Similarly, a comprehensive evaluation of different splitting criteria for ordinal trees was conducted in \cite{ayllon2024splitting}, identifying an ordinal adaptation of the Gini criterion as the most effective. Finally, Ghasemkhani et al. \cite{ghasemkhani2025ordinal} introduced the ordinal random tree with rank-oriented feature selection, which explicitly incorporates label ordering in both the feature selection and prediction stages, enhancing the model's ability to exploit the ordinal structure.

However, despite the existence of several approaches, there remains a subfield within OC that offers significant opportunities for the development of novel and effective methodologies: ordinal ensemble learning. Some of the methods proposed in this area include the work of Pérez-Ortíz et al. in \cite{perez2013projection}, introducing a methodology based on computing different classification tasks through the formulation of multiple order hypotheses (i.e., a reformulation of the well-known one-versus-all scheme). Fernández-Navarro et al. in \cite{fernandez2013negative} proposed two neural network threshold ensemble models for OC. The key difference between these models lies in the manner in which thresholds are determined: either as free parameters or fixed a priori, and whether these thresholds are allowed to be adapted during the training process. A more recent contribution by Guo et al. in \cite{guo2021ensemble} introduced an ensemble approach that integrates multi-binary classification, neuron stick-breaking, and soft label techniques, combining them through a median selection strategy. Similarly, Vega-Márquez et al. in \cite{vega2021ocean} proposed OCEAn, an ordinal ensemble framework in which the final prediction is determined by a weighted voting scheme. The weights are optimised using a genetic algorithm designed to minimise the classification cost. Vargas et al. in \cite{vargas2024ebano} presented an ordinal ensemble method in which individual models are aggregated using a novel weighting strategy based on a randomised search algorithm, allowing the ensemble to automatically assign relevance to each classifier.

Finally, over the last decade, several adaptations of AdaBoost to the OC setting have been proposed. All these approaches modify either the base classifier or the loss function used in the SAMME algorithm. For instance, in \cite{riccardi2014cost}, the authors proposed an AdaBoost approach that minimises a weighted least squares loss, using extreme learning machine as the base classifier. The main problem associated with this approach is that the incorporation of cost matrices inherently assumes a specific distance between class labels. Another ordinal variant of AdaBoost is introduced in \cite{singer2020ordinal}, where the classification accuracy used in the SAMME error function is replaced by an ordinal performance metric: a normalised version of the Mean Absolute Error (MAE). The MAE metric is the absolute deviation between predicted and target labels, considering the number of categories in the ordinal scale \cite{baccianella2009evaluation}. This again implies an assumption about the distance between categories, treating all adjacent labels as equally spaced. The most recent approach, known as Ordinal Encoding AdaBoost (OEAdaBoost) \cite{um2023adaptive}, combines the multiclass exponential loss function with a multi-dimensional encoding scheme for ordinal target variables. In addition, it uses the Multi-Layer Perceptron (MLP) as base classifiers, incorporating the ensemble loss in their optimisation process. The experiments performed in \cite{um2023adaptive} show that OEAdaBoost generally outperforms the cost-sensitive SAMME algorithm of \cite{riccardi2014cost} without assuming any distance between class labels. Moreover, OEAdaBoost has also been shown to outperform other well-known approaches, such as Random Forest (RF) \cite{breiman2001random} and a neural network model specifically adapted for OC (NNRank) \cite{cheng2008neural}.

In this work, \texttt{ADABORD} and its two key components are proposed: the decision tree classifier with Ordinal Gini (OGini) splitting criterion, which is used as base classifier, and the absolute Ranked Probability Score (aRPS), which is used as error function. Decision trees are the most commonly used base learners for boosting, particularly shallow trees, since boosting can combine such simple models in a highly adaptive and powerful manner \cite{hastie2009elements}. Moreover, since the aRPS error function is defined on cumulative probabilities, it does not require mapping class labels to arbitrary increasing real values; that is, it provides an ordinal-aware loss without imposing an artificial numerical scale on the classes. The iterative nature of AdaBoost allows the ensemble to correct the individual weaknesses of each tree, resulting in a highly accurate final classifier. In this way, additional performance improvements are expected with respect to the use of other approaches.

\section{Methodology} \label{sec:methodology}

In this section, the OC problem and the proposed \texttt{ADABORD} approach, along with its specific components, are defined.

\subsection{Problem definition} \label{subsec:problemdefinition}

The OC problem involves mapping a set of $K$ input variables (or features) to a discrete variable $\mathcal{Y}$ that satisfies an ordering constraint. The input associated to a pattern $i$ is commonly denoted as $\mathbf{x}_i \in \mathcal{X} \subseteq \mathbb{R}^K$, and the output label as $y_i\in\mathcal{Y} \in \{\mathcal{C}_1, \mathcal{C}_2, \ldots, \mathcal{C}_q, \ldots, \mathcal{C}_Q\}$, where $\mathcal{C}_1 \prec \mathcal{C}_2 \prec \ldots \prec \mathcal{C}_q \prec \ldots \prec \mathcal{C}_Q$. In this way, an OC dataset with $N$ patterns is given by $\mathcal{D} = \{(\mathbf{x}_1, y_1), (\mathbf{x}_2, y_2), \ldots, (\mathbf{x}_i, y_i), \ldots, (\mathbf{x}_N, y_N)\}$.

With respect to the output variable $\mathcal{Y}$, it is common to map the categories to a natural number using the function $v(\mathcal{C}_q) = q$. Although this transformation would assume a fixed distance between the ordinal categories, the distance is usually unknown and unequal, and different penalties should be associated with different classification errors. For instance, a misclassification error between two adjacent categories should be less penalised than an error between two distant categories in the ordinal scale.

\subsection{AdaBoost algorithm} \label{subsec:adaboost}
In this work, \texttt{ADABORD} is presented, which is built upon the nominal multiclass version of the AdaBoost algorithm, namely SAMME, originally introduced in \cite{zhu2009multi}. For the sake of completeness and self-containment, the SAMME algorithm is reproduced in Algorithm \ref{alg:samme} with minor changes to the notation.

\begin{algorithm}[!htpb]
            \caption{\textcolor{blue}{Multiclass SAMME (reproduced from \cite{zhu2009multi})}}
            \label{alg:samme}
            
            Initialise sample weights $w_i = 1/N$, \quad $i = 1, \ldots, N$\;
            
            \For{$m = 1$ \KwTo $M$}{
                Train a classifier $f_m(x)$ employing sample weights $w_i$\;
            
                Compute the misclassification error of $f_m(x)$ over the training set:
                \begin{equation}
                    \eacc = \frac{\sum_{i=1}^N w_i \cdot \mathds{1} \left( y_i \neq f_m(x_i) \right)}{\sum_{i=1}^N w_i}.
                    \label{eq:error_accuracy}
                \end{equation}
            
                Compute the weight associated to the estimator $f_m$:
                \begin{equation}
                    \alphacc = \log \left( \frac{1 - \eacc}{\eacc} \right) + \log(Q - 1).
                    \label{eq:alpha_computation}
                \end{equation}
            
                Update sample weights $w_i$:
                \begin{equation}
                    w_i \leftarrow w_i \cdot \exp\left( \alphacc \cdot \mathds{1} \left( y_i \neq f_m(\mathbf{x}_i) \right) \right), \quad i = 1, \ldots, N.
                    \label{eq:w_update}
                \end{equation}
            
                Re-normalise sample weights $w_i$:
                \begin{equation}
                    w_i = \frac{w_i}{\sum_{n=1}^N w_n}, \quad i = 1, \ldots, N.
                    \label{eq:renormalise}
                \end{equation}
            }
        
            Compute the final predictions using:
            \begin{equation}
                F(x_i) = \arg \max_q \sum_{m=1}^M \alphacc \mathds{1}(f_m(x_i) = q), \quad q = 1, \ldots, Q.
                \label{eq:final_pred}
            \end{equation}
            
            \end{algorithm}

At each iteration $m$, a weak classifier $f_m$ is trained in a weighted version of the dataset $\mathcal{D}_{(\mathbf{w})}$, where each pattern $i$, has an associated weight, $w_i$. These weights are referred to as sample weights, and conform a vector $\mathbf{w} \in \mathbb{R}^N$. In the AdaBoost setup, at iteration $m = 0$, sample weights all are equally set to $\frac{1}{N}$. At subsequent steps, sample weights are updated according to the error of the previous classifier (\Cref{eq:w_update}). This weights update performed at each iteration is known as boosting. The purpose of boosting is that, at each iteration, the estimator increases the focus on the patterns misclassified by previous estimators, so that the global additive model approximates to a Bayes classifier \cite{zhu2009multi}. Note that the error function considered in \Cref{eq:error_accuracy} is the inverse of the accuracy score. The term $\mathds{1} \left( y_i \neq f_m(x_i) \right)$ is equal to 1 if $y_i \neq f_m(x_i)$, and 0 otherwise.

In view of \Cref{eq:final_pred}, the AdaBoost method is equivalent to a weighted linear combination of $M$ weak classifiers trained sequentially. In this case, $\mathds{1}(f_m(x_i) = q)$ gives $1$ if $f_m(x_i) = q$, i.e. $f_m$ predicts class $\mathcal{C}_q$ for the pattern $x_i$, and $0$ otherwise. The value of $\alphacc$ (\Cref{eq:alpha_computation}) weights the prediction of $f_m$, and is calculated based on the error score obtained by that estimator (\Cref{eq:error_accuracy}) in the training set.

Below, the following two subsections describe the two key components of the \texttt{ADABORD} approach, namely, the base classifier and the error function, which, when embedded in the AdaBoost ensembling scheme, give rise to the proposed method, \texttt{ADABORD}.

\subsection{Base classifier: ordinal Decision Tree Classifier (DTC)} \label{subsec:odtc}

The first major contribution is the use of a DTC with OGini splitting criterion as the base estimator. Extensive experiments reported in \cite{ayllon2024splitting} demonstrate that OGini consistently outperforms other ordinal splitting criteria, making it the most effective choice for ordinal decision trees.

The OGini splitting criterion is specifically designed to fit the cumulative probability distribution of the output, making it particularly well-suited for its integration within the AdaBoost framework. As stated in \cite{piccarreta2008classification}, OGini accounts for the global distance between the conditional Cumulative Distribution Functions (CDFs) at each split. The OGini index of a tree node $D_m$ can be expressed as:
\begin{equation}
    \text{OGini}(D_m) = \sum_{q=1}^{Q}\hat{c}_{q|m} (1 - \hat{c}_{q|m}),
    \label{eq:ogini}
\end{equation}
where $\hat{c}_{q|m}$ is the cumulative frequency of observing class $q$ in the $D_m$. 

\subsection{Error function: absolute Ranked Probability Score (aRPS)} \label{subsec:rps}
The RPS, originally proposed in \cite{epstein1969scoring}, extends the Brier Score to accommodate ordinal or multi-categorical outcomes. Conceptually, it can be interpreted as the aggregate of Brier scores computed across all binary classification tasks generated by thresholding the ordinal scale at each pair of adjacent categories.
RPS quantifies the divergence between the predicted CDF and the empirical (or true) CDF. A lower RPS value indicates a more accurate probabilistic estimation, with the score reaching its minimum when the entire predictive probability mass is assigned to the observed category. This characteristic underscores the suitability of RPS for evaluating probabilistic estimations involving ordinal response variables.

In order to compute the RPS, the output variable $\mathcal{Y}$ is required to be encoded following a one-hot cumulative strategy. In this way, each $y_i$ will be transformed to a vector $\mathbf{c}_i \in \{0, 1\}^Q$ such that:
\begin{equation}
    c_{i,q} = \mathds{1}(y_i \leq q), \quad q = 1, 2, \ldots, Q.
\end{equation}

Similarly, the cumulative probabilities predicted by a model $f_m$ are denoted as $\hat{\mathbf{c}}_{m,i} \in \mathds{R}^Q$, such that $\hat{c}_{m,i,q}$ is the predicted cumulative probability of observing class $\mathcal{C}_q$ in the $i$-th pattern.

For the computation of the RPS, two main variants are distinguished. The first variant corresponds to the original RPS proposal, where the difference between predicted and observed cumulative distributions is squared, we will refer to this version as quadratic RPS (qRPS). The other variant, that is the one employed in \texttt{ADABORD}, replaces the square by the absolute error, and is named as absolute RPS (aRPS). In this way, the aRPS can be expressed as:
\begin{equation}
    \text{aRPS}(\mathbf{c}, \mathbf{\hat{c}}_m) = \frac{1}{N}\sum_{i=1}^N\sum_{q=1}^{\textcolor{blue}{Q-1}}|c_{i,q} - \hat{c}_{m,i,q}|,
    \label{eq:arps}
\end{equation}
where the last class ($\mathcal{C}_Q$) is not included since, under the standard cumulative encoding, both the target and predicted cumulative probabilities satisfy $c_{i,Q} = \hat{c}_{m,i,Q} = 1$, yielding a zero contribution. This aRPS is based on the proposal presented in \cite{galdran2023performance}, where the authors investigate the benefits of replacing the squared difference between real and predicted cumulative probabilities by the absolute difference to address certain problematic cases. 

It is worth mentioning that the aRPS leverages the ordinal nature of the categories without assuming or requiring any metric information regarding the distances between adjacent categories. Beyond its ability to evaluate the accuracy of probabilistic estimations, the aRPS offers a key advantage in the context of OC: it explicitly accounts for the inherent order among categories. Incorporating the aRPS as the loss function within an AdaBoost framework for OC offers several advantages. Unlike standard error metrics, the aRPS penalises misclassification errors based on the ordinal distance between predicted and true classes, encouraging classifiers to produce probability distributions that concentrate mass near the correct category. In this way, the aRPS encourages models to distribute probability in a manner that respects the ordinal structure. 

This characteristic makes the aRPS particularly suitable for the design and training of algorithms that not only aim to predict the correct class but also seek to minimise the ordinal distance between the predicted and actual outcomes. Such a property is especially valuable in applications where the severity of misclassification errors increases with the distance between categories, such as in credit risk assessment, medical diagnosis, or customer satisfaction prediction. Models optimised under the aRPS framework tend to generate predictions that better align with real-world cost functions, thereby enhancing both interpretability and decision-making.

The use of the aRPS in the context of AdaBoost allows for a more nuanced weighting of both weak learners and individual training instances: classifiers that make less severe ordinal errors receive higher weights, while instances misclassified with predictions far from the true class are emphasised more in subsequent iterations. As a result, the ensemble becomes more sensitive to the ordinal structure of the task, leading to improved predictive performance and better alignment with application-specific cost functions. In addition, the choice of aRPS over the standard RPS score enhances the training robustness, as it was observed that the original RPS produced a very low $\erps$, which led to a high $\alphrps$ and, consequently, a more dispersed distribution of the errors. A graphical comparison is provided in \Cref{subsec:understanding}, where this analysis is further extended by comparing the behaviour of both error functions, the original qRPS and the proposed aRPS.

Finally, it is worth noting that aRPS has two key advantages in limiting the influence of class imbalance \cite{altalhan2025imbalanced} and noisy or mislabelled instances \cite{shahri2022novel}. Unlike the nominal exponential loss used in the SAMME approach (Algorithm \ref{alg:samme}), which relies solely on predicted class labels, the proposed aRPS-based boosting framework evaluates the full predicted class distribution through an ordinal, distance-aware loss defined on cumulative probabilities. This formulation assigns graded penalties that increase with ordinal distance, thereby reducing the dominance of majority classes and increasing the influence of poorly modelled extreme minority instances during the reweighting process. In particular, errors in extreme (and typically under-represented) classes contribute non-zero deviations across multiple cumulative thresholds, increasing their relative influence compared to nominal losses based on a single label indicator. Moreover, the total loss is upper bounded by $Q-1$ (see \Cref{eq:arps}), independently of the predicted probabilities, and each term in the summation is bounded in $[0,1]$. In contrast to the exponential reweighting in SAMME, where misclassified instances receive weights that grow exponentially, the aRPS-based updates grow at most linearly with the cumulative deviation. By relying on absolute deviations, whose subgradients are bounded, noisy or mislabelled instances do not induce arbitrarily large updates, preventing them from dominating subsequent boosting iterations. As a consequence, aRPS mitigates the impact of class imbalance while simultaneously limiting sensitivity to noise, providing a closer surrogate to the ordinal absolute error, which helps explain the observed performance in challenging ordinal classification scenarios.

\subsection{\texttt{ADABORD}} \label{sec:adabord}
As noted above, the AdaBoost for ordinal classification (\texttt{ADA\-BORD}) algorithm proposed in this study (Algorithm \ref{alg:adabord}) introduces two substantial modifications to the standard AdaBoost scheme, i.e. SAMME presented in Algorithm \ref{alg:samme}, which have been detailed in previous subsections:
\begin{enumerate}
    \item A DTC with the Ordinal Gini splitting criterion is employed as the base estimator \cite{ayllon2024splitting} (\Cref{subsec:odtc}).
    \item The computation of the error $\eacc$ at each iteration is replaced by an error measure based on the aRPS, with the calculation of $\alphacc$ adapted accordingly (\Cref{subsec:rps}).
\end{enumerate}

\begin{algorithm}[htpb]
                \caption{\textcolor{blue}{\texttt{ADABORD}}}
                \label{alg:adabord}
                
                Initialise sample weights $w_i = 1/N$, $i = 1, 2, \ldots, N$\;
                
                \For{$m = 1$ \KwTo $M$}{
                    Train a DTC with OGini splitting criterion as base classifier $f_m(x)$, employing sample weights $w_i$\;
                
                    Compute the aRPS of $f_m(x)$ in the training set:
                    \begin{equation}
                        \erps = \frac{\text{\normalfont aRPS}(\mathbf{c}, \mathbf{\hat{c}}_m)}{Q - 1}
                        \label{eq:erps}
                    \end{equation}
                
                    Compute the weight associated to the estimator $f_m$:
                    \begin{equation}
                        \alphrps = \log \left( \frac{1 - \erps}{\erps} \right)
                        \label{eq:alphrps_computation}
                    \end{equation}
                
                    Update sample weights $w_i$:
                    \begin{equation}
                        w_i \leftarrow w_i \cdot \exp\left( \alphrps \cdot \mathds{1} \left( y_i \neq f_m(\mathbf{x}_i) \right) \right), \quad i = 1, \ldots, N.
                        \label{eq:w_update_adabord}
                    \end{equation}
                
                    Re-normalise sample weights $w_i$:
                    \begin{equation}
                        w_i = \frac{w_i}{\sum_{n=1}^N w_n}, \quad i = 1, \ldots, N.
                        \label{eq:renormalise_adabord}
                    \end{equation}
                }
                
                Compute final predictions:
                \begin{equation}
                    F(x_i) = \arg \max_q \sum_{m=1}^M \alphrps \mathds{1}(f_m(x_i) = q), \quad q = 1, \ldots, Q.
                    \label{eq:final_pred_adabord}
                \end{equation}
                
                \end{algorithm}

Note that the denominator $Q - 1$ included in the $\erps$ formula acts as normalisation term with respect to the number of classes. By adding this term, the computation of $\alphrps$ is reached, that is equal to $\alphacc$ but removing the $log(Q - 1)$, which is no longer required due to the aforementioned normalisation term.

It is worthy of mentioning that there is a synergy between the OGini criterion and the aRPS-based error, given that both components are designed to fit the cumulative probability distribution of the output, which contributes to improving the convergence of the algorithm in ordinal classification problems.

\begin{figure}[ht]
    \centering
    \includegraphics[width=.9\linewidth]{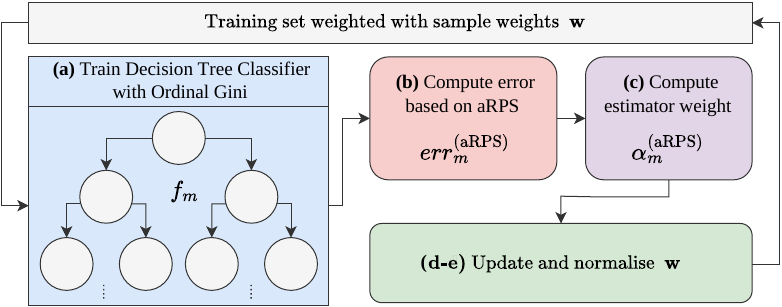}
    \caption{The \texttt{ADABORD} algorithm at each iteration $m$: (a) A decision tree classifier using ordinal Gini split criterion (\Cref{eq:ogini}) is trained. Each training instance is weighted with the corresponding sample weight contained in $\mathbf{w}$. This decision tree trained at iteration $m$ is denoted as $f_m$. (b) The error of $f_m$ is computed for the training set. The error score employed is based on the the absolute Ranked Probability Score (aRPS, \Cref{eq:arps}), as defined in \Cref{eq:erps}. (c) The weight $\alphrps$ (\Cref{eq:alphrps_computation}) of $f_m$ is calculated. (d-e) The sample weights $\mathbf{w}$ are updated based on the classification performance and the $\alphrps$ obtained by $f_m$ (\Cref{eq:w_update}). Finally, $\mathbf{w}$ is normalised so that the sum of all sample weights equals 1 \Cref{eq:renormalise}.}
    \label{fig:adabord_workflow}
\end{figure}

A graphical overview of the proposed \texttt{ADABORD} algorithm for a given iteration $m$ is provided in \Cref{fig:adabord_workflow}.

\section{Experimental setup} \label{sec:expsetup}

This section details the experimental protocol employed to validate the proposed \texttt{ADABORD} approach. First, the baseline methodologies used for comparative analysis are outlined; second, the benchmarking datasets employed in this study are described; and third, the performance metrics used to evaluate the models are defined. Finally, details of the experimental procedure are provided, including the cross-validation stage and the partitioning strategies, among others. Source code and instructions to reproduce the experiments are publicly available on the associated website\footnote{\url{https://www.uco.es/grupos/ayrna/adabord}}.

\subsection{Baseline techniques} \label{subsec:baseline}
In order to ensure a robust comparison with state-of-the-art methodologies, the following seven methodologies are considered:
\begin{itemize}
    \item AdaBoost \cite{zhu2009multi}. The standard multiclass AdaBoost method presented in \Cref{subsec:adaboost}, which uses a DTC as the base learner.
    \item Ordinal Encoding AdaBoost (OEAdaBoost) \cite{um2023adaptive}. OEAdaBoost represents the most recent adaptation of the standard AdaBoost framework to the OC paradigm. As previously detailed in \Cref{sec:related}, it combines a multiclass exponential loss function with a multidimensional encoding scheme. A notable characteristic of this approach is the use of Multi-Layer Perceptrons (MLPs) as base classifiers, where the ensemble loss is directly incorporated into the optimisation process. Among the three available ordinal AdaBoost variants \cite{singer2020ordinal,um2023adaptive,riccardi2014cost}, OEAdaBoost is selected as it has been shown to outperform cost-sensitive SAMME and other specific OC variants \cite{um2023adaptive}. Moreover, it is the only approach that does not rely on assuming arbitrary numerical values for the target labels.
    
    \item Back-Propagation Multi-Layer Perceptron (BP-MLP) \cite{bishop2006mlp}. A fully connected neural network architecture trained through the BP algorithm \cite{rumelhart1986learning}. In this regard, the adaptive moment estimation (Adam) \cite{diederik2015adam} is used as optimiser, and the cross-entropy is employed as loss function. The number of hidden neurons and layers is optimised through cross-validation.

    \item Back-Propagation Multi-Layer Perceptron with Cumulative Links Model output (BP-MLP-CLM) \cite{clm}. A fully connected neural network architecture that incorporates a cumulative link model (CLM) output layer, accounting for the ordinal structure of the target variable. The remaining hyperparameters, optimisation procedure, and loss function are the same as those used for BP-MLP.

    \item Ridge \cite{pedregosa2011scikit}. A ridge regressor adapted for multiclass classification by encoding the target variable into $\{-1, 1\}$ values, following a one-vs-all strategy for each class. The model learns linear decision functions by minimising a squared error loss with an $\ell_2$ regularization term. The regularization strength is controlled by a hyperparameter and is optimised via cross-validation.

    \item eXtreme Gradient Boosting (XGBoost) \cite{xgboost}. XGBoost is widely regarded as a gold standard for nominal classification. It represents an efficient and scalable implementation of gradient boosting machines that use Decision Tree Regressors (DTR) as base learners. XGBoost is well known for its strong predictive performance and regularisation capabilities.

    \item Ordinal Light Gradient Boosting Machine (O-LGBM) \cite{spanashis2024ordinalgbt}. A gradient boosting method for ordinal classification implemented on top of the LightGBM framework \cite{ke2017lightgbm}. It integrates a loss function tailored to ordinal targets. The underlying LightGBM framework relies on tree-based learning algorithms and is designed for high efficiency and scalability.
\end{itemize}

\subsection{Benchmarking datasets} \label{subsec:datasets}
\begin{figure}[ht!]
\centering
\begin{subfigure}{.49\linewidth}
\centering
  \includegraphics[width=0.8\linewidth]{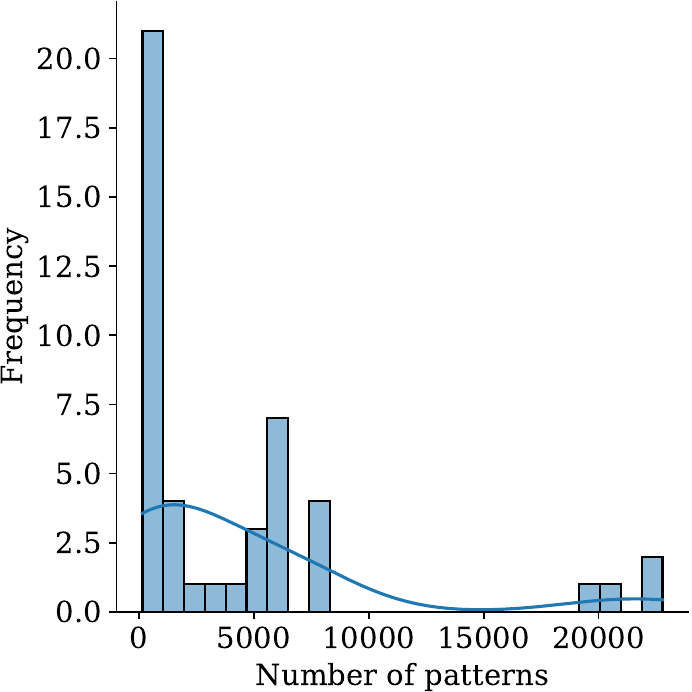}
  \caption{Distribution of the number of patterns ($N$).}
  \label{fig:tocuco_distrib_a}
\end{subfigure}
\begin{subfigure}{.49\linewidth}
\centering
  \includegraphics[width=0.8\linewidth]{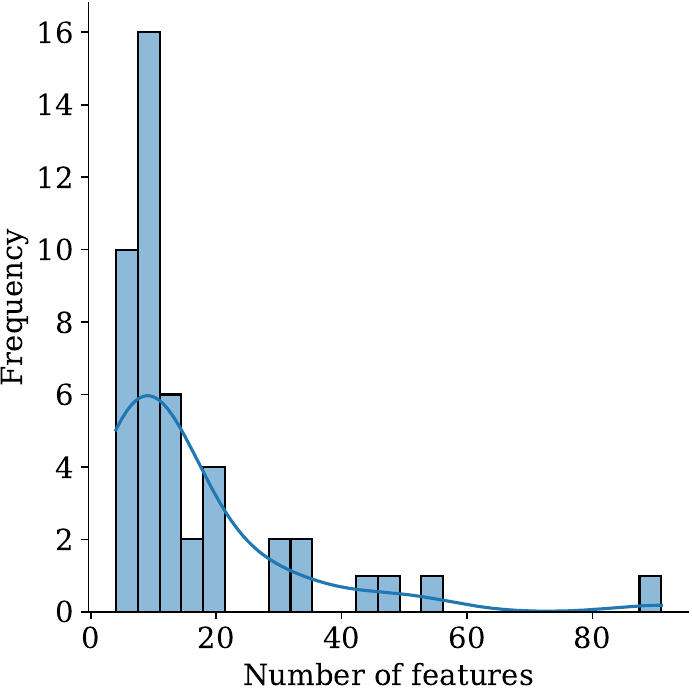}
  \caption{Distribution of the number of features ($K$).}
  \label{fig:tocuco_distrib_b}
\end{subfigure} \\[1cm]
\begin{subfigure}{.49\linewidth}
\centering
  \includegraphics[width=0.8\linewidth]{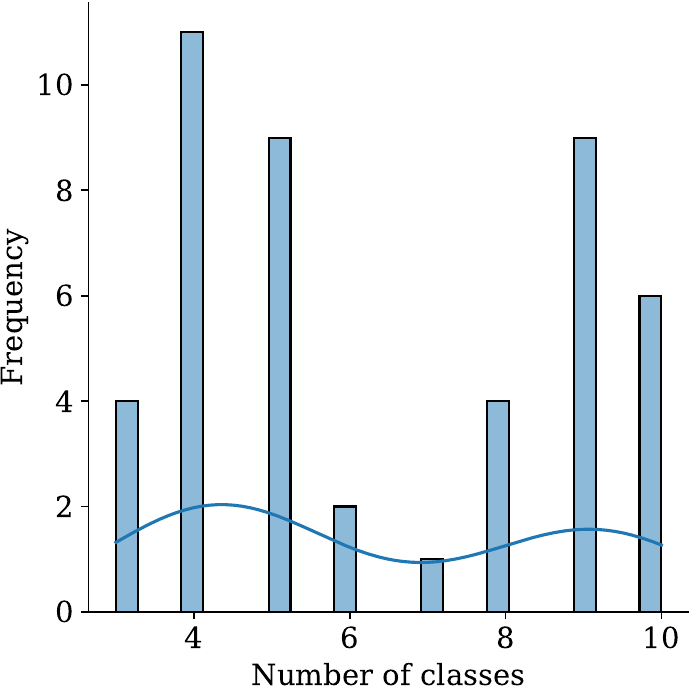}
  \caption{Distribution of the number of classes ($Q$).}
  \label{fig:tocuco_distrib_c}
\end{subfigure}
\begin{subfigure}{.49\linewidth}
\centering
  \includegraphics[width=0.8\linewidth]{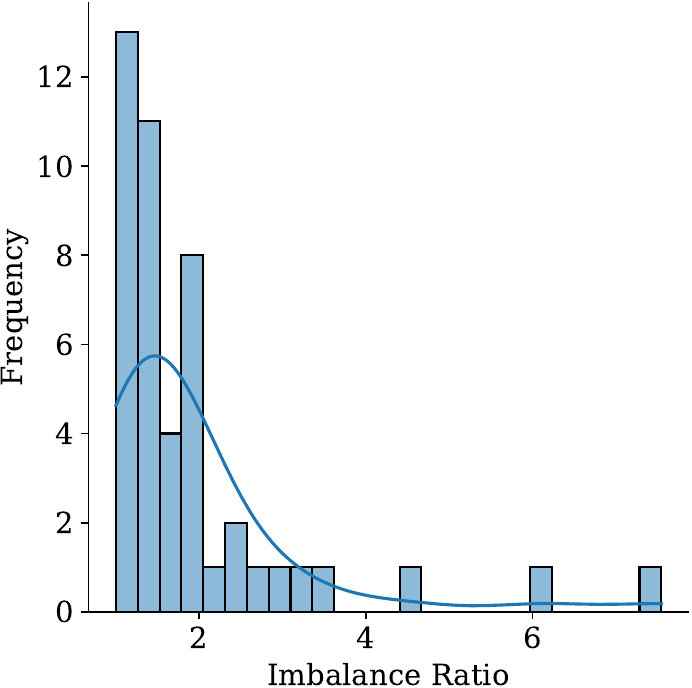}
  \caption{Distribution of the imbalance ratio.}
  \label{fig:tocuco_distrib_d}
\end{subfigure}
\caption{Distributions of the main characteristics of the TOC-UCO repository: (a) number of patterns ($N$), (b) number of features ($K$), (c) number of classes ($Q$), and (d) imbalance ratio.}
\label{fig:tocuco_distrib}
\end{figure}

In this study, the Tabular Ordinal Classification repository of the University of Córdoba (TOC-UCO) \cite{tocuco} is employed as benchmark. This repository contains a total of $46$ ordinal classification datasets. The characteristics of these datasets are presented in \Cref{tab:tocuco}, including the number of patterns in the train ($\#Train$) and test ($\#Test$) sets, the number of input features ($K$), the number of classes ($Q$), the distribution of patterns per class and the imbalanced ratio \cite{perez2016ordinalimbalance}. Among them, $24$ correspond to grouped continuous variables or discretised regression problems, where the output classes are derived through a discretisation process of an original regression variable. The remaining $22$ datasets involve inherently ordered categorical variables, where the ordinal classes are directly defined in the problem formulation. The distributions of the number of patterns, features, classes, and imbalance ratios within the TOC-UCO repository are illustrated in \Cref{fig:tocuco_distrib}. 
As shown in \Cref{fig:tocuco_distrib}, the TOC-UCO repository exhibits several characteristics that pose substantial challenges for state-of-the-art ordinal classifiers. On the one hand, there is a high amount of datasets with a small number of patterns ($N$) and input features ($K$), while on the other hand, some datasets contain considerably larger values (\Cref{fig:tocuco_distrib_a,fig:tocuco_distrib_b}). This heterogeneity makes it difficult to design models that are robust to overfitting in low-sample scenarios while remaining scalable to larger datasets. Moreover, the variability in the number of classes ($Q$), illustrated in \Cref{fig:tocuco_distrib_c}, has a significant impact on model performance. As $Q$ increases, ordinal constraints become more restrictive, often leading to a ``vanishing margin'' between adjacent classes. This phenomenon may arise from human subjectivity or annotation errors in originally ordinal datasets, or from instances lying near class boundaries when continuous targets are discretised using strict thresholds. Finally, the high imbalance ratios observed in \Cref{fig:tocuco_distrib_d} are also noteworthy. In ordinal settings, class imbalance is not only a matter of class frequency but also of class position, since minority classes located at the extremes of the ordinal scale can disproportionately bias overall performance.

\begin{table}[!htbp]
    \scriptsize
    \setlength{\tabcolsep}{2.5pt}
    \renewcommand{\arraystretch}{0.75}
    \caption{Characteristics of the \texttt{TOC-UCO} datasets. $K$ represents the number of input variables and $Q$ denotes the number of classes. The class distribution represents the percentage of patterns that belong to each class, so that the first value corresponds to class 1, second to class 2, etc. Imbalance ratio (IR) is computed based on Eq. (1) of \cite{tocuco}, and values equal to $1$ denote a perfectly balanced problem.}
    \label{tab:tocuco}
    \centering
    \begin{tabular}{lcccccc}
        \toprule \toprule
        \multicolumn{7}{c}{Discretised regression datasets (24)}\\
        \midrule
        Dataset                        & \#Train & \#Test & $K$ & $Q$ & Class distribution & IR \\
        \midrule
        forestfires              & 361                & 156               & 8           & 4          & (0.75 0.13 0.04 0.07)                                  & 3.51 \\
        machine                  & 146                & 63                & 6           & 4          & (0.70 0.14 0.08 0.08)                                  & 2.44 \\
        buoysFlux46026           & 4090               & 1754              & 8           & 5          & (0.39 0.24 0.15 0.11 0.10)                             & 1.36 \\
        buoysFlux46059           & 4090               & 1754              & 8           & 5          & (0.45 0.25 0.13 0.08 0.09)                             & 1.62 \\
        census1                  & 15948              & 6836              & 8           & 5          & (0.46 0.24 0.13 0.08 0.09)                             & 1.63 \\
        census2                  & 15948              & 6836              & 16          & 5          & (0.46 0.24 0.13 0.08 0.09)                             & 1.63 \\
        buoysFlux46069           & 4090               & 1754              & 8           & 6          & (0.34 0.24 0.16 0.10 0.07 0.09)                        & 1.42 \\
        calhousing               & 14448              & 6192              & 8           & 7          & (0.24 0.19 0.16 0.14 0.10 0.08 0.09)                   & 1.18 \\
        buoysHeight46026         & 4090               & 1754              & 8           & 8          & (0.20 0.16 0.16 0.12 0.11 0.09 0.06 0.10)              & 1.14 \\
        buoysHeight46069         & 4090               & 1754              & 8           & 8          & (0.21 0.18 0.14 0.13 0.11 0.09 0.05 0.09)              & 1.20 \\
        cancerTreatment          & 748                & 321               & 12          & 8          & (0.31 0.14 0.11 0.15 0.09 0.06 0.07 0.07)              & 1.31 \\
        abalone                  & 2923               & 1254              & 10          & 9          & (0.20 0.14 0.16 0.15 0.12 0.06 0.05 0.03 0.09)         & 1.46 \\
        bank1                    & 5734               & 2458              & 8           & 9          & (0.29 0.14 0.12 0.10 0.08 0.07 0.06 0.05 0.09)         & 1.31 \\
        buoysHeight46059         & 4090               & 1754              & 8           & 9          & (0.20 0.16 0.14 0.12 0.10 0.08 0.07 0.05 0.09)         & 1.20 \\
        computer1                & 5734               & 2458              & 12          & 9          & (0.07 0.02 0.03 0.06 0.08 0.11 0.17 0.21 0.25)         & 1.97 \\
        computer2                & 5734               & 2458              & 21          & 9          & (0.07 0.02 0.03 0.06 0.08 0.11 0.17 0.21 0.25)         & 1.97 \\
        housing                  & 354                & 152               & 13          & 9          & (0.18 0.12 0.19 0.17 0.12 0.07 0.05 0.04 0.06)         & 1.37 \\
        insurance                & 936                & 402               & 9           & 9          & (0.28 0.19 0.16 0.12 0.06 0.05 0.03 0.03 0.08)         & 1.75 \\
        soybean                  & 224                & 96                & 9           & 9          & (0.21 0.18 0.13 0.12 0.07 0.07 0.08 0.05 0.08)         & 1.24 \\
        bank2                    & 5734               & 2458              & 32          & 10         & (0.45 0.13 0.09 0.07 0.05 0.04 0.03 0.03 0.03 0.07)    & 2.04 \\
        cancerDeathRate          & 2132               & 915               & 29          & 10         & (0.10 0.09 0.10 0.11 0.12 0.13 0.11 0.08 0.06 0.10)    & 1.05 \\
        concreteStrength         & 721                & 309               & 8           & 10         & (0.14 0.08 0.13 0.14 0.10 0.11 0.08 0.07 0.06 0.09)    & 1.08 \\
        realState                & 289                & 125               & 6           & 10         & (0.10 0.09 0.10 0.09 0.11 0.13 0.11 0.09 0.07 0.11)    & 1.03 \\
        stock                    & 665                & 285               & 9           & 10         & (0.14 0.08 0.06 0.12 0.11 0.10 0.09 0.08 0.08 0.13)    & 1.08 \\
        \midrule
        \multicolumn{7}{c}{Originally OC datasets (22)}\\
        \midrule
        Dataset                        & \#Train & \#Test & K & Q & Class distribution & IR \\
        \midrule
        balanceScale             & 437                & 188               & 4           & 3          & (0.46 0.08 0.46)                                    & 2.35 \\
        mammoexp                 & 288                & 124               & 5           & 3          & (0.57 0.25 0.18)                                    & 1.38 \\
        newthyroid               & 150                & 65                & 5           & 3          & (0.14 0.70 0.16)                                    & 1.96 \\
        tae                      & 105                & 46                & 54          & 3          & (0.32 0.33 0.34)                                    & 1.00 \\
        car                      & 1209               & 519               & 21          & 4          & (0.70 0.22 0.04 0.04)                               & 4.46 \\
        childrenAnemia           & 4098               & 1757              & 14          & 4          & (0.28 0.28 0.41 0.03)                               & 2.88 \\
        gymExerciseTracking      & 681                & 292               & 17          & 4          & (0.20  0.38 0.31 0.10)                              & 1.36 \\
        heartDisease             & 205                & 89                & 13          & 4          & (0.64 0.13 0.09 0.15)                               & 1.97 \\
        LESTSensors              & 3578               & 1534              & 6           & 4          & (0.16 0.14 0.23 0.46)                               & 1.30 \\
        LEVXSensors              & 3578               & 1534              & 6           & 4          & (0.33 0.08 0.10 0.48)                               & 1.93 \\
        problematicInternet      & 1484               & 636               & 32          & 4          & (0.59 0.27 0.13 0.01)                               & 7.55 \\
        support                  & 511                & 219               & 19          & 4          & (0.39 0.13 0.08 0.40)                               & 1.79 \\
        swd                      & 700                & 300               & 10          & 4          & (0.03 0.35 0.40 0.22)                               & 3.10 \\
        eucalyptus               & 515                & 221               & 91          & 5          & (0.24 0.15 0.18 0.29 0.14)                          & 1.10 \\
        lev                      & 700                & 300               & 4           & 5          & (0.09 0.28 0.40 0.20 0.03)                          & 2.70 \\
        nhanes                   & 3656               & 1567              & 30          & 5          & (0.11 0.28 0.40 0.18 0.04)                          & 2.15 \\
        vlbw                     & 120                & 52                & 19          & 5          & (0.15 0.17 0.15 0.28 0.25)                          & 1.10 \\
        winequalityRed           & 1119               & 480               & 11          & 5          & (0.04 0.43 0.40 0.12 0.01)                          & 6.11 \\
        esl                      & 341                & 147               & 4           & 6          & (0.11 0.20 0.24 0.28 0.13 0.05)                     & 1.51 \\
        studentPerformance       & 463                & 199               & 43          & 8          & (0.07 0.04 0.11 0.24 0.22 0.16 0.11 0.06)           & 1.50 \\
        era                      & 700                & 300               & 4           & 9          & (0.09 0.14 0.18 0.17 0.16 0.12 0.09 0.03 0.02)      & 1.86 \\
        melbourneAirbnb          & 14025              & 6011              & 48          & 10         & (0.08 0.09 0.10 0.09 0.05 0.12 0.11 0.09 0.09 0.18) & 1.12 \\
        \bottomrule \bottomrule
    \end{tabular}
\end{table}

\subsection{Performance metrics} \label{subsec:metrics}
For evaluating the different methodologies, the following four performance metrics are considered: three ordinal metrics that explicitly account for the ordered nature of the classes, and one standard (nominal) metric that considers the imbalance degree of the problems. Regarding the ordinal performance metrics, although most of them originate from the regression context, they are presented below in their ordinal classification formulation, as previously detailed in \cite{CRUZRAMIREZ201421}:
\begin{itemize}
    \item Averaged Mean Absolute Error (AMAE) \cite{baccianella2009evaluation}. This metric computes the individual Mean Absolute Error ($\text{MAE}_q$) for each class $\mathcal{C}_q$:
    \begin{equation} \label{eq:mae}
        \text{MAE}_q = \frac{1}{N_q} \sum_{i \mid y_i \in \mathcal{C}_{q}} \left| v(y_i) - v(\widehat{y_i}) \right| \quad \in [0, Q-1],
    \end{equation} where $N_q$ denotes the number of patterns belonging to class $\mathcal{C}_q$. The AMAE is then obtained by averaging the individual MAEs across all classes:
    \begin{equation} \label{eq:amae}
        \text{AMAE} = \frac{1}{Q} \sum_{q=1}^Q \text{MAE}_q, \quad \in [0, Q-1].
    \end{equation}
    This metric must be minimised and is particularly suitable for assessing performance under class imbalance.

    \item Quadratic Weighted Kappa (QWK) \cite{delatorreQWK2018}. This is a maximisation metric that extends Cohen's Kappa index \cite{cohen1968weighted} by incorporating a penalty term that increases quadratically with the distance between predicted and true classes \cite{warrens2012cohen}. It is defined as:
    \begin{equation}
        \text{QWK} = 1 - \frac{\sum\limits_{q=1}^Q \sum\limits_{j=1}^Q \omega_{qj} O_{qj}}{\sum\limits_{q=1}^Q \sum\limits_{j=1}^Q \omega_{qj} E_{qj}}, \quad \in [-1, 1],
    \end{equation}
    where $\omega_{qj} = \frac{(q-j)^2}{(Q-1)^2} \in [0,1]$ is the $(q,j)$-th entry of the penalty matrix (reflecting the quadratic cost of misclassification), 
    $O_{qj}$ represents the $(q,j)$-th element of the confusion matrix, and $E_{qj} = \frac{O_{q\bullet} O_{\bullet j}}{N}$ denotes the expected count under the assumption of random predictions. Here, $O_{q\bullet}$ and $O_{\bullet j}$ are the row and column sums of the confusion matrix, respectively.
    
    \item Maximum Mean Absolute Error (MMAE) \cite{CRUZRAMIREZ201421}. Similar to AMAE, but instead of averaging the class-specific $\text{MAE}_q$, it considers the maximum error among all classes:
    \begin{equation}
        \text{MMAE} = \max_{q=1}^Q \left\{ \text{MAE}_q \right\}, \quad \in [0, Q-1].
    \end{equation}
    Like AMAE, MMAE is a minimisation metric and highlights worst-case performance, which is particularly relevant in imbalanced scenarios.
    
    \item Balanced Accuracy (BACC) \cite{brodersen2010balanced}. This maximisation metric measures the average sensitivity (recall) across all classes. Unlike standard accuracy, BACC mitigates the influence of class imbalance by assigning equal weight to each class, regardless of its frequency in the dataset. It is defined as:
\begin{equation}
    \text{BACC} = \frac{1}{Q} \sum_{j=1}^Q \frac{O_{jj}}{O_{j\bullet}}, \quad \in [0, 1],
\end{equation}
where $O_{jj}$ denotes the diagonal elements of the confusion matrix (i.e., correctly classified instances for class $\mathcal{C}_j$), and $O_{j\bullet}$ represents the total number of samples in class $\mathcal{C}_j$ (the sum of the $j$-th row of $O$).
\end{itemize}

AMAE, QWK, and MMAE are considered as appropriate performance metrics as they take into account the ordinality and the imbalance of the problem, two properties inherent in all the datasets employed. The BACC is also considered to provide a non-ordinal perspective on the performance of the models.

\subsection{Data partitioning and cross-validation strategy} \label{subsec:partitioning}
In order to enhance reproducibility, the partitions provided in the TOC-UCO repository \cite{tocuco} are considered. Each dataset is partitioned into train and test sets following a $70-30\%$ stratified hold-out strategy. These partitions are resampled $30$ times using different random seeds, so that for each dataset $30$ different train-test splits are obtained.

Each method included in the experiments is cross-validated on each training partition by means of a randomised search with $20$ iterations. At each iteration, a hyperparameter configuration is evaluated using a stratified $3$-fold strategy. For each model, the best configuration is selected according to the AMAE performance metric (\Cref{eq:amae}). The hyperparameter values considered for each method are presented in \Cref{tab:crossvalidation}. It is worth noting that the number of estimators (\#Estimators) is set differently for \texttt{ADABORD} and AdaBoost compared to OEAdaBoost, in order to prevent overfitting in the former case. As previously mentioned, while OEAdaBoost employs the MLP as base classifier, both \texttt{ADABORD} and AdaBoost consider DT-based classifiers. This rationale is also observed in the original OEAdaBoost proposal \cite{um2023adaptive}, where the \#Estimators is higher for DT-based classifiers.
\begin{table}[ht]
    \centering
    \setlength{\tabcolsep}{3pt}
    \caption{\textcolor{blue}{Experimental setup of the different methods, including the training algorithm employed, the base estimator, and the corresponding hyperparameter configurations. Hyperparameters shown as sets indicate the candidate values considered during cross-validation.}}
    \label{tab:crossvalidation}
    \resizebox{\textwidth}{!}{
    \begin{tabular}{llll}
    \toprule\toprule
    Method(s) & Algorithm & Base estimator & Hyperparameter \& Values \\
    \midrule

    \multirow{3}{*}{\texttt{ADABORD}} & AdaBoost & DTC
        & \multirow{3}{*}{\#Estimators: \{$50$, $100$, $250$, $500$, $1000$, $2000$\}} \\
    & \quad + aRPS loss & \quad + OGini splitting
        & \\[0.1cm]
    \multirow{2}{*}{AdaBoost} & AdaBoost & DTC
        & \multirow{2}{*}{Estimator max depth: \{$4$, $8$, $16$, $\text{None}^1$\}}\\
    & \quad + Multiclass exponential loss & \quad + Gini splitting
        & \\[0.1cm]
    \midrule

    \multirow{2}{*}{OEAdaBoost} & AdaBoost & \multirow{2}{*}{MLP}
        & \#Estimators: \{$10$, $25$, $50$, $100$, $200$\} \\
    & \quad + Multiclass exponential loss &
        & Estimator \#hidden layers: \{$2$, $4$, $6$, $8$\} \\[0.1cm]
    \midrule

    \multirow{4}{*}{BP-MLP} & \multirow{4}{*}{Back-propagation} & \multirow{4}{*}{MLP}
        & Epochs: \{$100$, $250$, $500$, $1000$\} \\
    & & & \#Hidden layers: \{$1$, $2$, $4$\} \\
    & & & \#Hidden neurons: \{$4$, $8$, $16$\} \\
    \multirow{5}{*}{BP-MLP-CLM} & \multirow{5}{*}{Back-propagation} & \multirow{5}{*}{MLP} & Learning rate: $10^{-3}$ \\
    & & \multirow{5}{*}{\quad + CLM output layer} & Learning rate scheduler: Reduce on plateau$^2$ \\
    & & & Optimiser: Adam \\
    & & & Activation function: ReLU \\
    & & & Minimum distance$^3$: \{$0.0$, $0.1$, $0.2$\} \\[0.1cm]
    \midrule
    Ridge & Least squares & Linear regressor
        & Regularisation strength: \{$10^{-3}$, $10^{-2}$, \ldots, $10^3$\} \\[0.1cm]
    \midrule
    \multirow{2}{*}{XGBoost} & \multirow{2}{*}{XGBoost} & \multirow{2}{*}{DTR} & \#Estimators: \{$100$, $250$, $500$, $1000$\} \\
    & \multirow{2}{*}{\quad + Regularised Taylor-expanded} & \multirow{2}{*}{\quad + Gradient-hessian} & Learning rate: \{$1\times10^{-2}$, $5\times10^{-2}$, $1\times10^{-1}$\} \\
    & \multirow{1.5}{*}{\quad \quad loss} & \multirow{1.5}{*}{\quad \quad gain splitting} & Pattern subsample \%: \{$0.75$, $0.95$, $1.0$\} \\
    \multirow{1.5}{*}{O-LGBM} & \multirow{1.5}{*}{LightGBM} & \multirow{1.5}{*}{DTR} & Max depth: \{$3$, $5$, $8$\} \\
    & \multirow{1.5}{*}{\quad + ordinal regression loss} & \multirow{1.5}{*}{\quad + leaf-wise histograms} & Column subsample \%: \{$0.75$, $0.95$, $1.0$\} \\[0.1cm]
    
    \bottomrule\bottomrule
    \multicolumn{4}{l}{$^1$ The tree grows without considering any depth limit.}\\
    \multicolumn{4}{l}{$^2$ Employing a factor $= 0.75$, patience $=50$, minimum learning rate $= 10^{-5}$, and improvement threshold $= 10^{-2}$.}\\
    \multicolumn{4}{l}{$^3$ Only used in BP-MLP-CLM to adjust the minimum threshold considered in class separation.}\\
    \end{tabular}
    }
\end{table}

\section{Results and discussion} \label{sec:results}

\subsection{Experimental results} \label{subsec:exp_results}

This section presents the results obtained under the experimental setup described in the previous section, in which \texttt{ADABORD} is compared against seven state-of-the-art techniques, including OEAdaBoost \cite{um2023adaptive}, which, to the best of the authors' knowledge is the state-of-the-art AdaBoost-based ensemble method for OC. The comparison is conducted using four performance metrics. In addition to analysing the complete set of $46$ datasets, those datasets with $5$ or more classes are separately examined (i.e., a subset of $31$ problems from the TOC-UCO repository). The rationale for this distinction is to better evaluate the impact of ordinality: as the number of classes increases, the ordinal nature of the problem becomes more pronounced, and the performance gap between ordinal and nominal methods becomes more evident. Consequently, a statistical study is conducted on this subset to rigorously assess the differences between the proposed \texttt{ADABORD} and the current state-of-the-art approaches. All individual results are publicly available on the associated website\footnote{\url{https://www.uco.es/grupos/ayrna/adabord}.}

In the left part of \Cref{tab:means_stds}, the mean and standard deviation (STD) of each method, computed by averaging over the $46$ datasets and the $30$ executions, are presented for the different performance metrics. Notably, no single method consistently outperforms the others across all metrics. In terms of AMAE and MMAE, the proposed \texttt{ADABORD} approach achieves the best results, with OEAdaBoost obtaining the second best performance. Regarding the QWK metric, XGBoost and OEAdaBoost rank as the best and second-best approaches, respectively. Finally, in terms of BACC, XGBoost achieves the highest result, followed by \texttt{ADABORD}.

When the analysis is restricted to datasets with $5$ or more classes (a total of $31$ datasets), where the ordinal nature of the problem is more pronounced, it can be observed (right part of \Cref{tab:means_stds}) that \texttt{ADABORD} achieves the best performance in terms of AMAE, QWK, and MMAE. XGBoost performs very closely in terms of QWK and obtains the best results for BACC. Remarkably, AdaBoost achieves the second-best results for AMAE, MMAE, and BACC. This behaviour highlights that capturing ordinality is not a straightforward task and that, in some cases, nominal approaches can still serve as competitive baselines.

\begin{sidewaystable}[htpb!]
    \centering
    \caption{Mean and standard deviation (STD) presented as Mean$\text{STD}$ for AMAE, QWK, MMAE and BACC. Metrics to minimise are followed by ($\downarrow$), and metrics to maximise with ($\uparrow$). Best values are highlighted in bold, and second best in italics. The results for all (46) datasets and for those datasets with $5$ or more classes ($Q \geq 5$, 31 out of the 46 datasets), are shown.}
    \label{tab:means_stds}
    \begin{tabular}{lcccc|cccc}
        \toprule\toprule
        & \multicolumn{4}{c|}{All datasets (46)} & \multicolumn{4}{c}{Datasets with $Q \geq 5$ (31 out of 46)} \\
        \cmidrule(lr){2-5} \cmidrule(lr){6-9}
        Method & AMAE ($\downarrow$) & QWK ($\uparrow$) & MMAE ($\downarrow$) & BACC ($\uparrow$)
               & AMAE ($\downarrow$) & QWK ($\uparrow$) & MMAE ($\downarrow$) & BACC ($\uparrow$) \\
        \midrule
        AdaBoost     & $0.758_{0.047}$           & $0.735_{0.026}$           & $1.268_{0.145}$           & $0.519_{0.025}$ 
                     & $\mathit{0.854_{0.046}}$ & $0.785_{0.017}$           & $\mathit{1.418_{0.154}}$ & $\mathit{0.475_{0.021}}$ \\
        OEAdaBoost   & $\mathit{0.747_{0.044}}$ & $\mathit{0.741_{0.023}}$  & $\mathit{1.258_{0.130}}$  & $0.502_{0.025}$ 
                     & $0.858_{0.043}$          & $0.782_{0.017}$           & $1.424_{0.139}$           & $0.446_{0.022}$ \\
        \texttt{ADABORD} & $\mathbf{0.734_{0.047}}$  & $0.740_{0.025}$           & $\mathbf{1.214_{0.137}}$  & $\mathit{0.520_{0.027}}$ 
                     & $\mathbf{0.823_{0.046}}$ & $\mathbf{0.790_{0.018}}$  & $\mathbf{1.342_{0.147}}$  & $0.474_{0.024}$ \\       
        BP-MLP          & $0.908_{0.148}$           & $0.677_{0.070}$           & $1.467_{0.342}$           & $0.477_{0.051}$ 
                     & $1.116_{0.195}$          & $0.698_{0.079}$           & $1.809_{0.452}$           & $0.398_{0.057}$ \\
        BP-MLP-CLM          & $0.960_{0.191}$           & $0.681_{0.076}$           & $1.540_{0.411}$           & $0.469_{0.054}$ 
                     & $1.172_{0.253}$          & $0.702_{0.091}$           & $1.883_{0.541}$           & $0.391_{0.064}$ \\
        Ridge        & $1.059_{0.055}$           & $0.645_{0.028}$           & $1.765_{0.121}$           & $0.424_{0.025}$ 
                     & $1.295_{0.057}$          & $0.675_{0.020}$           & $2.166_{0.133}$           & $0.350_{0.020}$ \\
        XGBoost      & $0.761_{0.046}$           & $\mathbf{0.743_{0.025}}$  & $1.295_{0.138}$           & $\mathbf{0.523_{0.023}}$ 
                     & $0.860_{0.045}$          & $\mathit{0.789_{0.017}}$  & $1.439_{0.148}$           & $\mathbf{0.479_{0.020}}$ \\
        O-LGBM        & $0.889_{0.059}$           & $0.724_{0.028}$           & $1.590_{0.134}$           & $0.452_{0.020}$ 
                     & $0.998_{0.069}$          & $0.779_{0.023}$           & $1.772_{0.160}$           & $0.399_{0.022}$ \\
        \bottomrule\bottomrule
    \end{tabular}
\end{sidewaystable}

In \Cref{tab:rankings}, the results are presented in terms of rankings. The proposed \texttt{ADABORD} method achieves the best (lowest) rank for AMAE and MMAE, and the second-best rank for BACC when considering the entire TOC-UCO repository. Conversely, the OEAdaBoost method obtains the best rank in QWK and the second-best ranks in AMAE and MMAE. Remarkably, XGBoost stands out in terms of BACC, and achieving the second-best rank for QWK.

Similarly to \Cref{tab:means_stds}, the right part of \Cref{tab:rankings} shows the results for datasets with $5$ or more classes. It can be observed that \texttt{ADABORD} outperforms the other techniques in AMAE, QWK, and MMAE, while XGBoost achieves the best rank in BACC (second-best in terms of this performance metric is for \texttt{ADABORD}). Notably, OEAdaBoost achieves the second-best ranks for AMAE, MMAE, and QWK.

\begin{table}[htpb]
    \centering
    \caption{Ranking results for the different performance metrics. Best (lowest) values are highlighted in bold, while second best are highlighted in italics. Results for all (46) datasets and for those datasets with 5 or more classes ($Q \geq 5$, 31 out of the 46 datasets) are included}.
    \label{tab:rankings}
    \resizebox{\textwidth}{!}{
    \begin{tabular}{lcccc|cccc}
        \toprule\toprule
        \multirow{2.3}{*}{Method} & \multicolumn{4}{c|}{All datasets (46)} & \multicolumn{4}{c}{Datasets with $Q \geq 5$ (31 out of 46)} \\
        \cmidrule(lr){2-5} \cmidrule(lr){6-9}
          & AMAE   & QWK    & MMAE   & BACC   
          & AMAE   & QWK    & MMAE   & BACC \\
        \midrule
        AdaBoost        & $3.837$ & $4.054$ & $3.772$ & $3.772$ & $3.226$ & $3.742$ & $3.387$ & $3.290$ \\
        OEAdaBoost       & $\mathit{3.326}$ & $\mathbf{3.087}$ & $\mathit{3.522}$ & $4.087$ & $\mathit{3.065}$ & $\mathit{2.968}$ & $\mathit{3.290}$ & $4.032$ \\
        \texttt{ADABORD}    & $\mathbf{2.902}$ & $3.207$ & $\mathbf{2.880}$ & $\mathit{3.554}$ & $\mathbf{2.065}$ & $\mathbf{2.548}$ & $\mathbf{2.194}$ & $\mathit{3.097}$ \\
        BP-MLP& $4.913$ & $6.043$ & $4.522$ & $4.891$ & $5.968$ & $6.677$ & $5.419$ & $5.581$ \\
        BP-MLP-CLM    & $5.543$ & $5.652$ & $5.022$ & $5.348$ & $6.226$ & $6.161$ & $5.613$ & $5.806$ \\
        Ridge      & $6.435$ & $6.630$ & $5.978$ & $5.565$ & $7.452$ & $7.161$ & $6.710$ & $6.323$ \\
        XGBoost        & $3.652$ & $\mathit{3.109}$ & $4.326$ & $\mathbf{3.087}$ & $3.161$ & $3.000$ & $3.806$ & $\mathbf{2.581}$ \\
        O-LGBM     & $5.391$ & $4.217$ & $5.978$ & $5.696$ & $4.839$ & $3.742$ & $5.581$ & $5.290$ \\
        \bottomrule\bottomrule
    \end{tabular}}
\end{table}

\subsection{Comparative analysis} \label{subsec:analysis}

\begin{figure}[h!]
\centering
\begin{subfigure}{.45\linewidth}
  \includegraphics[width=\linewidth]{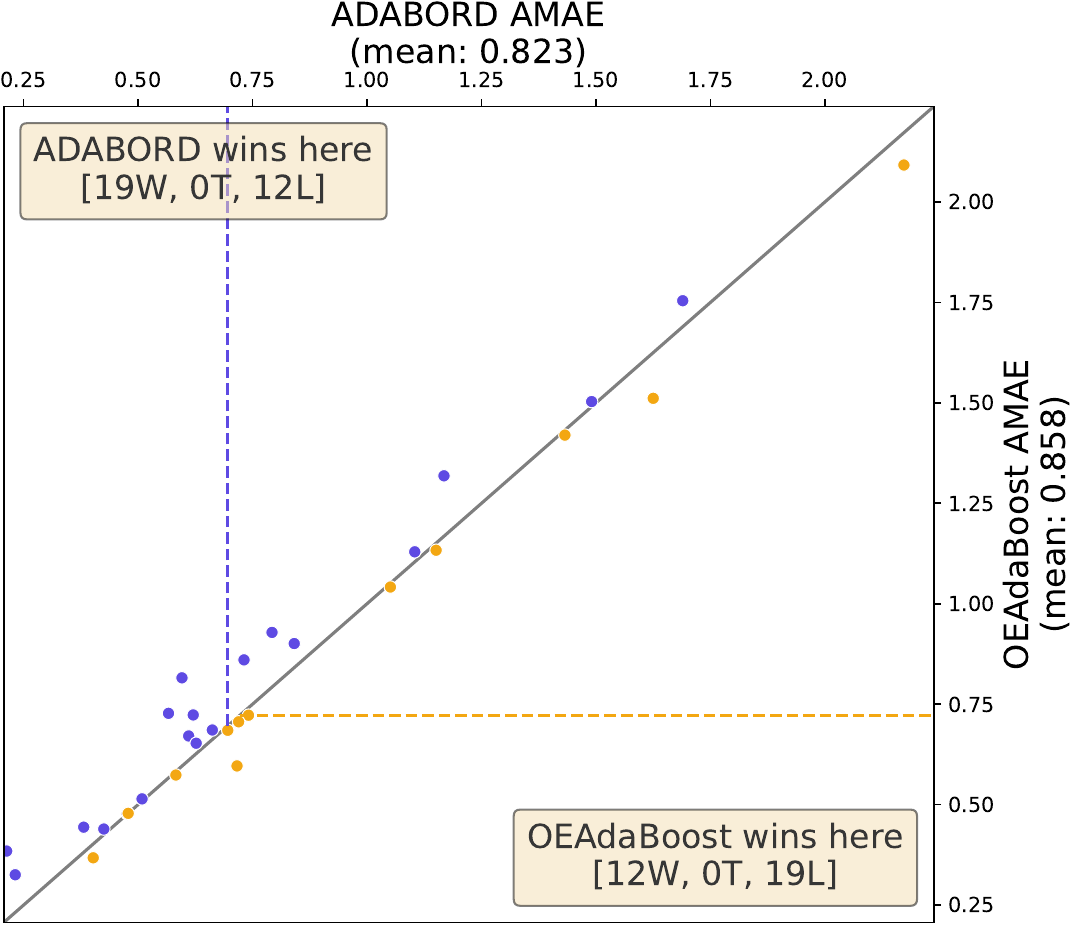}
  \caption{AMAE}
  \label{fig:scatter_AMAE}
\end{subfigure}
\begin{subfigure}{.45\linewidth}
  \includegraphics[width=\linewidth]{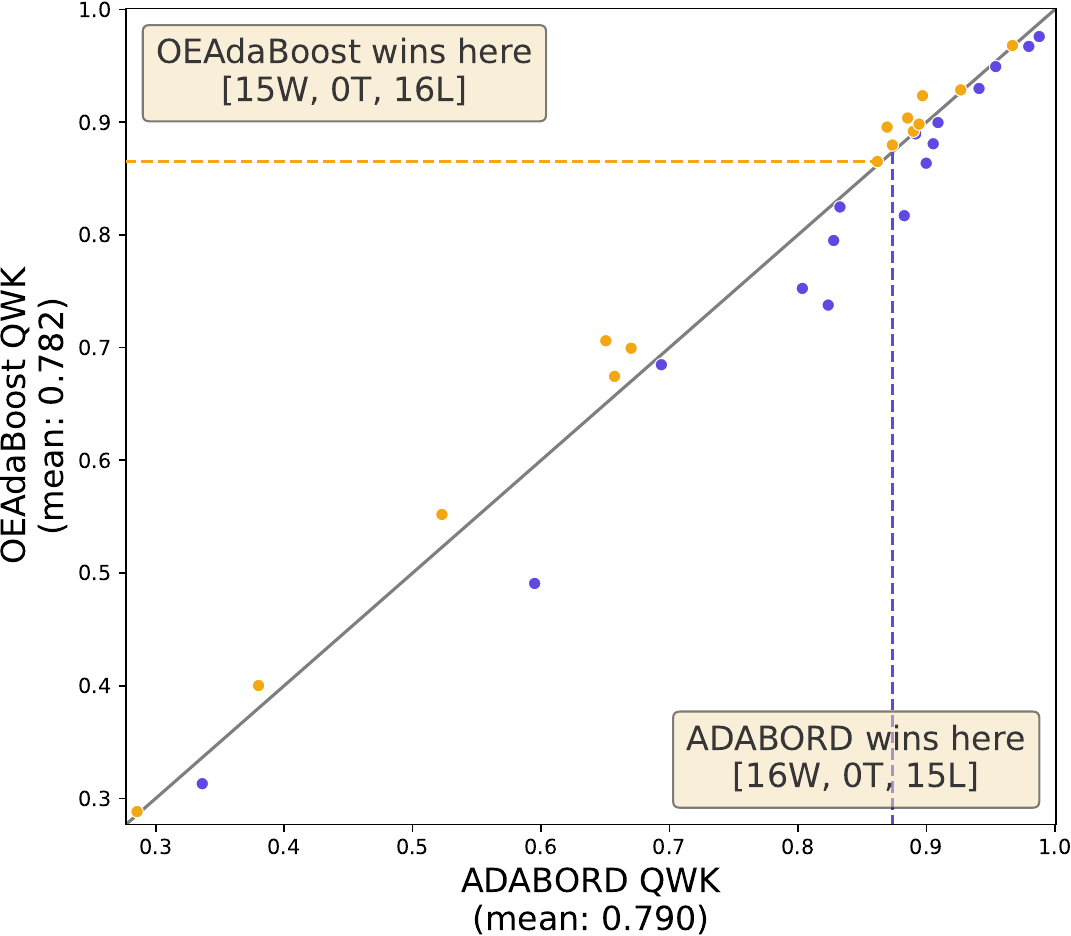}
  \caption{QWK}
  \label{fig:scatter_QWK}
\end{subfigure}\\[1cm]
\begin{subfigure}{.45\linewidth}
  \includegraphics[width=\linewidth]{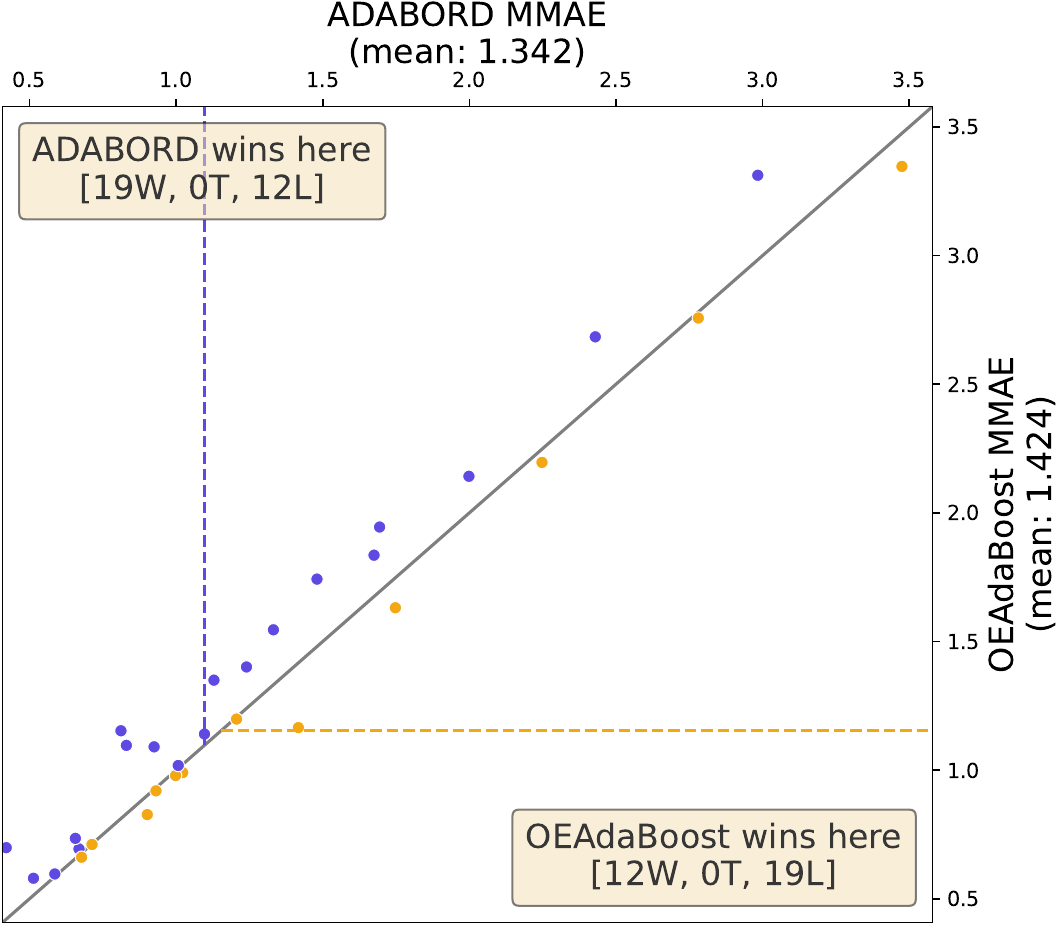}
  \caption{MMAE}
  \label{fig:scatter_MMAE}
\end{subfigure}
\begin{subfigure}{.45\linewidth}
  \includegraphics[width=\linewidth]{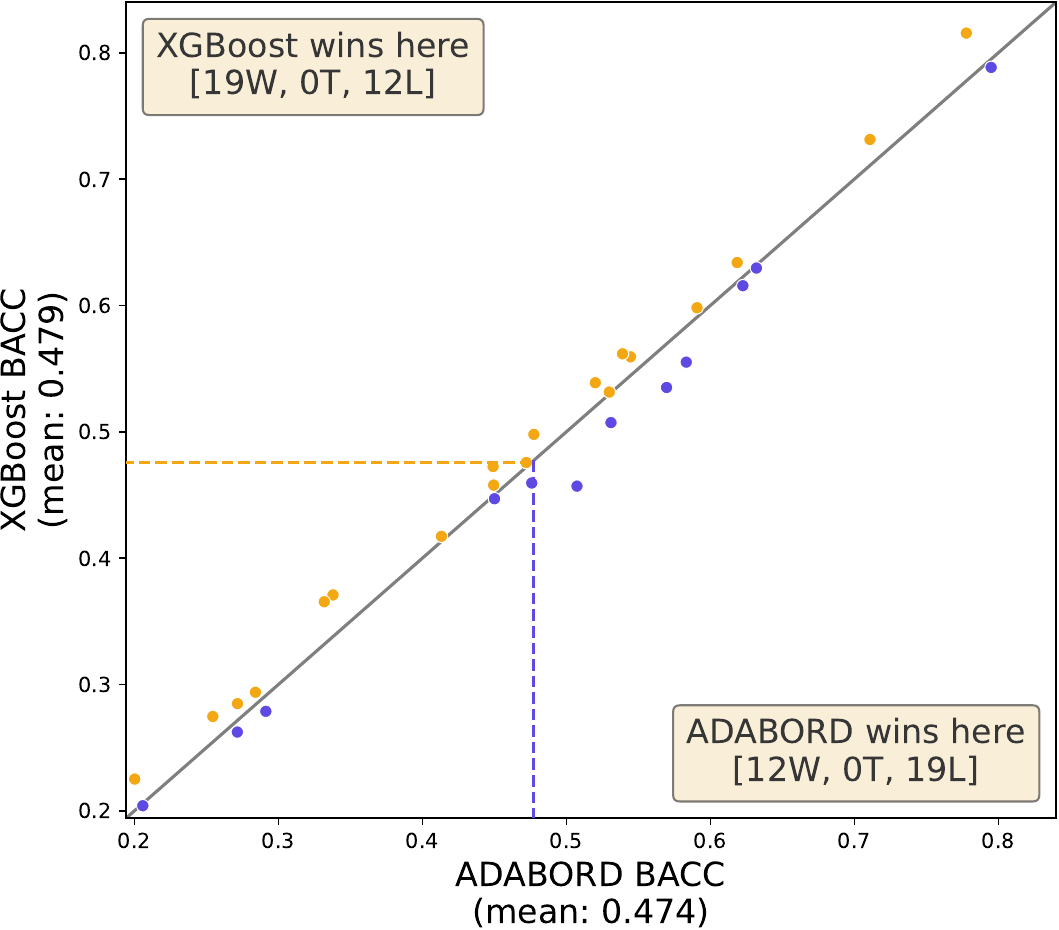}
  \caption{BACC}
  \label{fig:scatter_BACC}
\end{subfigure}
\caption{Scatter plots comparing the two best-performing approaches for each evaluation metric: (a) \texttt{ADABORD} versus OEAdaBoost in terms of AMAE, (b) \texttt{ADABORD} versus OEAdaBoost in terms of QWK, (c) \texttt{ADABORD} versus OEAdaBoost in terms of MMAE, and (d) \texttt{ADABORD} versus XGBoost in terms of BACC. For metrics to be minimised (AMAE and MMAE), the best-performing method is placed on the y-axis, whereas for metrics to be maximised (QWK and BACC), the best-performing method is placed on the x-axis. The competing method is shown on the opposite axis. Each point corresponds to the result obtained on a single dataset. Points located on the \texttt{ADABORD} side (coloured in blue) indicate results favourable to this method. Dashed lines denote the median value obtained for each method--metric pair. The top-right and bottom-left corners display the number of wins (W), ties (T), and losses (L) for each method. The code used to generate these plots was adapted from \cite{middlehurst2024aeon}.}
\label{fig:scatter}
\end{figure}

To graphically analyse the differences between the two top-performing approaches for each performance metric, scatter plots showing pairwise com\-pa\-ri\-sons are displayed in \Cref{fig:scatter}. Since the objective is to better evaluate the results on datasets with more pronounced ordinality, the analysis is restricted to those datasets with $5$ or more classes (a total of $31$ out of the $46$). For AMAE (\Cref{fig:scatter_AMAE}), QWK (\Cref{fig:scatter_QWK}), and MMAE (\Cref{fig:scatter_MMAE}), the majority of points lie on the \texttt{ADABORD} side (blue colour), indicating a superior performance of this method compared to OEAdaBoost (orange colour). However, regarding the nominal metric BACC (\Cref{fig:scatter_BACC}), although the mean values are very close, XGBoost (orange colour) outperforms \texttt{ADABORD} (blue colour) in terms of the number of wins ($19$ vs. $12$, respectively).

\begin{figure}[htpb]
    \centering
    \includegraphics[width=0.9\linewidth]{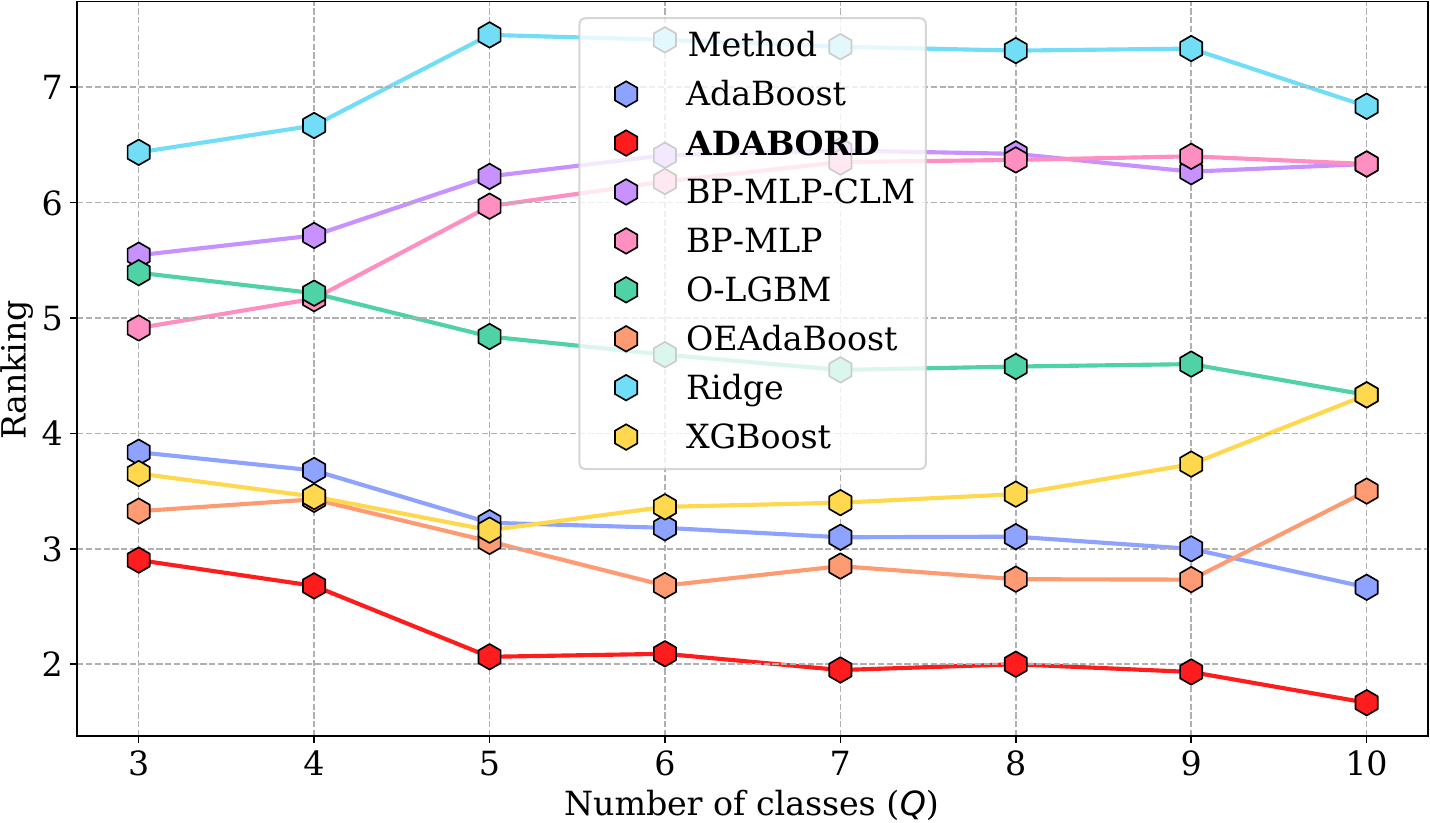}
    \caption{Ranks achieved, in terms of AMAE, by the different methodologies for each number of classes ($Q$). These ranks are computed as the average of the individual ranks obtained on all datasets with exactly $Q$ classes.}
    \label{fig:all_methods_rank_by_n_classes}
\end{figure}

Moreover, in \Cref{fig:all_methods_rank_by_n_classes}, the evolution of the rankings obtained by each method across different numbers of classes is analysed. To this end, the datasets are grouped by their number of classes and the average ranking per group is computed using the AMAE metric. It can be observed that the proposed \texttt{ADABORD} method achieves the best performance in every group. Additionally, its rankings tend to improve as the number of classes increases, indicating that \texttt{ADABORD} performs particularly well when ordinal constraints become more restrictive, i.e., the ordinal classification task becomes more challenging. In contrast, the rankings of XGBoost progressively worsen, and it is even outperformed by the nominal AdaBoost method starting from the group with $5$ classes, where the ordinal nature of the problems becomes more pronounced. Regarding OEAdaBoost, it remains the second-best performer across all groups, except in the group with the highest number of classes ($10$), where it is surpassed by AdaBoost.

\begin{figure}[htpb]
    \centering
    \includegraphics[width=0.9\linewidth]{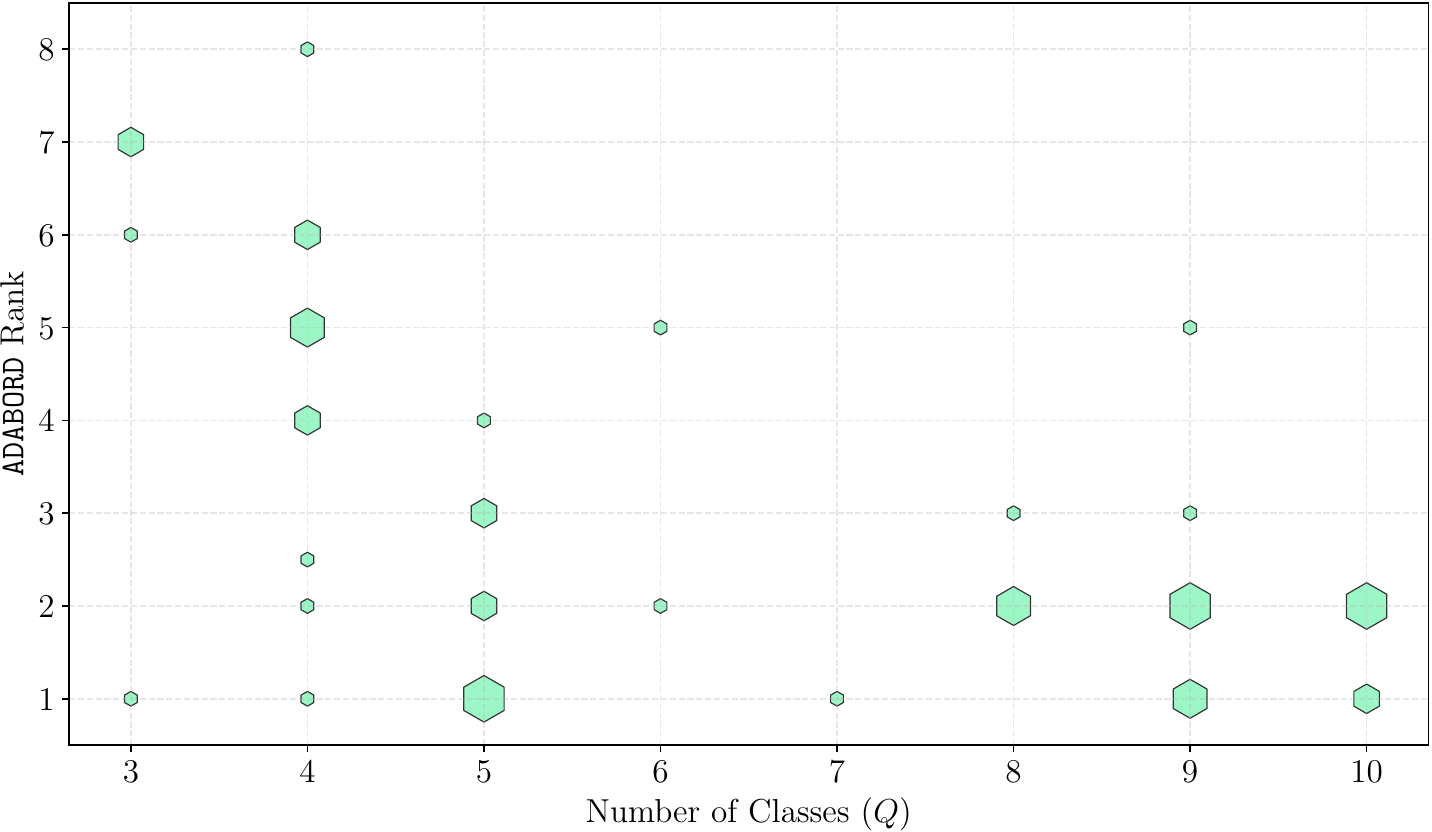}
    \caption{Evolution of the ranks obtained by the \texttt{ADABORD} approach for each number of classes ($Q$), according to the AMAE performance metric. The size of each point is proportional to the number of datasets in which \texttt{ADABORD} achieved that specific rank, with the smallest and largest points representing $1$ and $4$ datasets, respectively.}
    \label{fig:adabord_ranking_across_n_classes}
\end{figure}

In \Cref{fig:adabord_ranking_across_n_classes}, the evolution of the ranking of the \texttt{ADABORD} method as the number of classes increases is examined. In this case, instead of averaging different datasets, the rank obtained by \texttt{ADABORD} in every individual dataset is shown, grouped by the number of classes. The size of each point is proportional to the number of datasets in which the same rank is achieved. Notably, the rankings tend to decrease (i.e., improve) as the number of classes increases. Focusing on datasets with $5$ or more classes (a total of $31$), \texttt{ADABORD} achieves either the first or second rank in $25$ datasets, with $10$ in first place and $14$ in second. Moreover, in the groups with $7$, $8$, $9$, and $10$ classes, \texttt{ADABORD} achieves either the first or second rank in all but three datasets.

In summary, as shown in \Cref{fig:all_methods_rank_by_n_classes,fig:adabord_ranking_across_n_classes}, the performance gap favouring \texttt{ADABORD} tends to grow with the number of classes, highlighting its strength in more complex OC tasks.

\subsection{Computational complexity analysis}

Regarding computational complexity, \Cref{tab:compcomplexity} reports the complexity associated with the methods compared in this work.
        \begin{table}[ht]
        \centering
        \setlength{\tabcolsep}{3pt}
        \caption{\textcolor{blue}{Comparison of the computational complexity of each approach evaluated in this work.}}
        \label{tab:compcomplexity}
        \resizebox{0.95\textwidth}{!}{
        \begin{tabular}{llll}
        \toprule\toprule
        Method & Algorithm & Base estimator  & Computational complexity \\
        \midrule
        \multirow{2}{*}{AdaBoost} & AdaBoost & DTC & \multirow{2}{*}{$\mathcal{O}(M \cdot N \cdot K^2)$} \\
        &\quad + Multiclass exponential loss& \quad + Gini splitting & \\ [0.25cm]
        \multirow{2}{*}{OEAdaBoost} & AdaBoost & \multirow{2}{*}{MLP}  &  \multirow{2}{*}{$\mathcal{O}(M \cdot E \cdot L \cdot N \cdot K^2)$} \\
        & \quad + Multiclass exponential loss & & \\[0.25cm]
        \multirow{2}{*}{\texttt{ADABORD}} & AdaBoost & DTC & \multirow{2}{*}{$\mathcal{O}(M \cdot N \cdot K^2)$} \\
        & \quad + aRPS loss & \quad + OGini splitting  & \\[0.25cm]
        BP-MLP  & Back-propagation  & MLP  &    $\mathcal{O}(E \cdot L \cdot N \cdot K^2)$ \\[0.25cm]
        \multirow{2}{*}{BP-MLP-CLM}  & \multirow{2}{*}{Back-propagation}  & MLP &    \multirow{2}{*}{$\mathcal{O}(E \cdot L \cdot N \cdot K^2)$} \\
        & & \quad + CLM output layer & \\[0.25cm]
        Ridge & Least squares & Linear regressor  &  $\mathcal{O}(N \cdot K^2)$    \\[0.25cm]
        \multirow{2}{*}{XGBoost} & XGBoost & DTR  & \multirow{2}{*}{$\mathcal{O}(M \cdot N \cdot d  \cdot K \cdot \log K)$} \\
        & \quad + Regularised Taylor-expanded loss & \quad + Gradient-hessian gain splitting & \\[0.25cm]
        \multirow{2}{*}{O-LGBM} & LightGBM & DTR  & \multirow{2}{*}{$\mathcal{O}(M \cdot N \cdot d \cdot K \cdot \log N)$}\\
        & \quad + Ordinal immediate thresholds loss & \quad + Histogram-based gradient gain splitting & \\
        \bottomrule\bottomrule
        \multicolumn{4}{l}{$M$ is the number of estimators, $N$ is the number of training samples, $K$ is the number of input variables, $E$ is the number of}\\
        \multicolumn{4}{l}{epochs, $L$ is the number of layers, and $d$ is the depth of the trees.} \\
        \end{tabular}
        }
        \end{table}

As can be observed from \Cref{tab:compcomplexity}, the original AdaBoost algorithm has a complexity of $\mathcal{O}(M \cdot N \cdot K^2)$ \cite{hu2008adaboost}, where $M$ denotes the number of estimators, $N$ the number of training samples, and $K$ the number of input variables. The term $N \cdot K^2$ corresponds to the complexity of the base learner, which in the original AdaBoost formulation is the standard DTC. The proposed \texttt{ADABORD} method differs from the nominal AdaBoost approach in two main aspects: the use of an ordinal base estimator (the standard DTC using the Ordinal Gini splitting criterion) and the adoption of the aRPS as the error function. The overall boosting framework remains unchanged. Therefore, the computational complexity of \texttt{ADABORD} is equivalent to that of the original AdaBoost algorithm.

OEAdaBoost extends the AdaBoost framework by employing an MLP as base learner instead of a decision tree. Consequently, its complexity becomes $\mathcal{O}(M \cdot E \cdot L \cdot N \cdot K^2)$, where $E$ denotes the number of training epochs and $L$ the number of hidden layers, reflecting the iterative cost of training an MLP at each epoch. BP-MLP and BP-MLP-CLM, which are trained through back-propagation, have complexity $\mathcal{O}(E \cdot L \cdot N \cdot K^2)$, which depends on the number of epochs and network depth. Unlike OEAdaBoost, the MLP is trained only once, rather than repeatedly within an ensemble procedure, resulting in a lower computational cost than its boosted counterpart. Ridge regression presents the lowest computational burden among the compared methods, with complexity $\mathcal{O}(N \cdot K^2)$, since model fitting is based on solving a regularised least-squares problem. XGBoost has computational complexity $\mathcal{O}(M \cdot d \cdot N \cdot K \cdot \log K)$, where $d$ is the maximum tree depth. This formulation arises from the sequential construction of regularised regression trees using first- and second-order gradient information. Although more sophisticated than standard boosting trees, XGBoost is generally efficient in practice due to its optimised implementation and pruning strategies. Finally, the O-LGBM has a similar computational complexity to XGBoost with a key difference: O-LGBM uses an histogram-based splitting instead of sorting all values as XGBoost.

\subsection{Statistical analysis} \label{subsec:statistical}

In this section, the statistical analysis of the results is described. This study is restricted to datasets with $5$ or more classes, as this allows for drawing more robust conclusions regarding the ordinality of the methods. A total of $248$ experiments have been conducted ($31$ datasets and eight methodologies), and for each experiment, $30$ executions have been performed using different random seeds. Consequently, for each performance measure, $248$ samples are obtained, each of size $30$. With the goal of applying parametric tests to compare the average results obtained for each performance metric, a non-parametric Kolmogorov-Smirnov test (K-S test) with $\alpha = 0.050$ has been applied to assess the normality of each sample. The results of these tests indicate that all $248$ samples follow a normal distribution, and therefore it is appropriate to conduct parametric tests. Hence, a two-factor ANalysis Of VAriance (ANOVA II) test is applied for each performance metric, considering both the dataset and the methodology, as well as their interaction.

For instance, the linear model used to compute the AMAE mean based on the two factors and their interaction can be expressed as:

\begin{equation}
\begin{aligned}
    \text{AMAE}_{i,j,k} &= \mu + D_i + M_j + (DM)_{i,j} + \varepsilon_{i,j,k},\\
    & i = 1, \ldots, 31; \quad j = 1, \ldots, 8; \quad k = 1, \ldots, 30,
\end{aligned}
\end{equation}
where $\mu$ denotes the fixed effect common to all populations, $D_i$ represents the effect of the $i$-th dataset, and $M_j$ denotes the effect of the $j$-th methodology. The term $(DM)_{i,j}$ captures the interaction between the $i$-th dataset and the $j$-th methodology, while $\varepsilon_{i,j,k}$ accounts for residual effects, including random variations. Here, $\text{AMAE}_{i,j,k}$ is the response variable used in the analysis. Analogous models are constructed for the remaining three performance metrics (MMAE, QWK, and BACC).

The ANOVA II results (for reference, the results corresponding to the AMAE metric are reported in \Cref{tab:anova_amae}) show that, for all the considered metrics, both the dataset and the methodology have a statistically significant influence on the mean performance values. This reveals that: 1) across different datasets, the same methodology exhibits statistically significant differences in the mean values of the four metrics; 2) for a fixed dataset, the eight methodologies also present significant differences in their mean values for all metrics; and 3) the interaction between dataset and methodology significantly impacts the mean values of each metric considered.

\begin{table}[!ht]
\centering
\caption{ANOVA II results for the AMAE metric. For each factor and interaction, the Degrees of Freedom (DF), Sum of Squares (SS), Mean Squares (MS), F-score, and significance level ($p$-value) are reported. Factors with $p < 0.050$ are considered statistically significant at the $95\%$ confidence level.}
\label{tab:anova_amae}
\resizebox{0.8\textwidth}{!}{
\begin{tabular}{lccccc}
\toprule\toprule
Source & SS & DF & MS & F-score & p-value \\
\midrule
Corrected model & 2004.691 & 247 & 8.116 & 413.199 & $<$0.001 \\
Intersection & 7395.894 & 1 & 7395.894 & 376529.413 & $<$0.001 \\
Dataset & 1589.716 & 30 & 52.991 & 2697.781 & $<$0.001 \\
Methodology & 207.429 & 7 & 29.633 & 1508.620 & $<$0.001 \\
Dataset $\times$ Methodology & 207.545 & 210 & 0.988 & 50.316 & $<$0.001 \\
Error & 141.267 & 7440 & 0.020 &  &  \\
\midrule
Total & 9541.852 & 7440 &  &  &  \\
Corrected total & 2145.958 & 7439 &  &  &  \\
\bottomrule\bottomrule
\end{tabular}}
\end{table}

Based on the ANOVA II results, a Tukey post-hoc test is performed to assess whether there are significant differences between individual methodologies. The purpose of Tukey's HSD multiple comparison test is to group the eight methodologies into clusters of statistically similar performance, such that differences between clusters are statistically significant at the $\alpha = 0.050$ level, while no significant differences are observed, on average, within each cluster. \Cref{tab:tukey_amae} presents the results of the Tukey test for the AMAE metric, showing that the proposed \texttt{ADABORD} method is significantly superior to the remaining techniques. The second-best methodology is AdaBoost, with no significant difference compared to OEAdaBoost and XGBoost (p-value $=0.987$). Regarding the QWK metric (\Cref{tab:tukey_qwk}), the best methodologies are \texttt{ADABORD}, XGBoost, and AdaBoost which exhibit similar performance (p-value $=0.300$). \texttt{ADABORD} significantly outperforms the other techniques. As for MMAE (\Cref{tab:tukey_mmae}), \texttt{ADABORD} significantly surpasses all other methods. Notably, the second-best group consists of standard AdaBoost, OEAdaBoost, and XGBoost. Finally, for the BACC metric (\Cref{tab:tukey_bacc}), XGBoost achieves the best results, but it is not significantly superior to AdaBoost or \texttt{ADABORD} (p-value $=0.265$).

\begin{table}[ht!]
\centering
\caption{Tukey test results in terms of the four metrics considered in this work: (a) AMAE, (b) QWK, (c) MMAE, and (d) BACC. The results are limited to those datasets with $5$ or more classes.}

    \begin{subtable}{0.51\textwidth}
        \centering
        \caption{AMAE}
        \resizebox{\textwidth}{!}{
        \begin{tabular}{lcccccc}
        \toprule \toprule
        & \multicolumn{6}{c}{Group Rank AMAE ($\downarrow$)} \\
        \cmidrule{2-7}
        Method          & 1         & 2         & 3     & 4  & 5 & 6    \\
        \midrule
        \texttt{ADABORD}            & $0.823$   &         &     &    & &     \\
        AdaBoost             &       & $0.854$     &    &     & &    \\
        OEAdaBoost            &        & $0.858$     &      &  & &    \\ 
        XGBoost             & & $0.860$     &     &    & &   \\
        O-LGBM             & &      & $0.998$    &    & &   \\
        BP-MLP             &     &       &  & $1.116$    & &   \\
        BP-MLP-CLM             &     &       & & & $1.172$     &   \\
        Ridge           &       &      &     &  & &  $1.295$ \\
        \midrule
        p-values        & 1.000     & $0.987$     & 1.000 & 1.000 & 1.000 &  1.000\\
        \bottomrule \bottomrule
    \end{tabular}}
    \label{tab:tukey_amae}
    \end{subtable}
    \hspace{0.27cm}
    \begin{subtable}{0.45\textwidth}
        \centering
        \caption{QWK}
        \resizebox{\textwidth}{!}{
        \begin{tabular}{lccccc}
        \toprule \toprule
        & \multicolumn{5}{c}{Group Rank QWK ($\uparrow$)} \\
        \cmidrule{2-6}
        Method      & 1         & 2             & 3         & 4  & 5   \\
        \midrule
        \texttt{ADABORD}        & $0.790$       &          &       &  &  \\
        XGBoost         & $0.789$       &  $0.789$         &        &  &   \\
        AdaBoost         & $0.785$    &   $0.785$        &   $0.785$   &       & \\
        OEAdaBoost        &   &   $0.782$         &  $0.782$ &      & \\
        O-LGBM        &   &            & $0.779$  &      & \\
        BP-MLP-CLM         &     &       &        & $0.702$ & \\ 
        BP-MLP         &     &       &        & $0.698$ & \\ 
        Ridge       &    &        &      &      & $0.675$ \\
        \midrule
        p-values        &$0.300$      &   $0.109$   & $0.347$ & $0.740$ & $1.000$ \\
        \bottomrule \bottomrule
        \end{tabular}}
    \label{tab:tukey_qwk}
    \end{subtable}\\[0.4cm]
    \begin{subtable}{0.48\textwidth}
        \centering
        \caption{MMAE}
        \resizebox{\textwidth}{!}{
        \begin{tabular}{lccccc}
        \toprule \toprule
        & \multicolumn{5}{c}{Group Rank MMAE ($\downarrow$)} \\
        \cmidrule{2-6}
        Method          & 1         & 2         & 3     & 4   & 5  \\
        \midrule
        \texttt{ADABORD}       & $1.342$   &       &     &      &       \\
        AdaBoost        &  & $1.418$    &     &      &     \\
        OEAdaBoost       &     & $1.424$    &   &    & \\
        XGBoost        &    &  $1.439$    &    & &       \\
        O-LGBM        &    &     & $1.773$   & &       \\
        BP-MLP        &       &         & $1.809$  &   &      \\
        BP-MLP-CLM        &       &         &   & $1.883$  &      \\ 
        Ridge      &      &      &   &      & $2.166$   \\
        \midrule
        p-values        & $1.000$     & $0.865$     & $0.225$ & $1.000$ & $1.000$ \\
        \bottomrule \bottomrule
    \end{tabular}}
    \label{tab:tukey_mmae}
    \end{subtable}
    \hspace{0.27cm}
    \begin{subtable}{0.48\textwidth}
        \centering
        \caption{BACC}
        \resizebox{\textwidth}{!}{
        \begin{tabular}{lccccc}
        \toprule \toprule
        & \multicolumn{5}{c}{Group Rank BACC ($\uparrow$)} \\
        \cmidrule{2-6}
        Method          & 1         & 2     & 3  & 4 & 5   \\
        \midrule
        XGBoost         & $0.479$ &      &             & & \\ 
        AdaBoost         &   $0.475$     &        &     & &  \\
        \texttt{ADABORD}        &    $0.474$    &     & &   &  \\
        OEAdaBoost        &  &  $0.446$   & &     &       \\ 
        O-LGBM         &   &   & $0.399$   &  & \\
        BP-MLP         &   &   & $0.398$   &  & \\
        BP-MLP-CLM         &     & & & $0.391$     & \\
        Ridge       &  &       &  &  &   $0.350$  \\        
        \midrule
        p-values        &  $0.265$   & $1.000$     &  $0.998$ &$1.000$ &  $1.000$\\
        \bottomrule \bottomrule
    \end{tabular}}
    \label{tab:tukey_bacc}
    \end{subtable}
\end{table}

\subsection{Understanding the error function in \texttt{ADABORD}} \label{subsec:understanding}

\begin{figure}[hbt!]
\centering
\begin{subfigure}{.49\linewidth}
  \includegraphics[width=\linewidth]{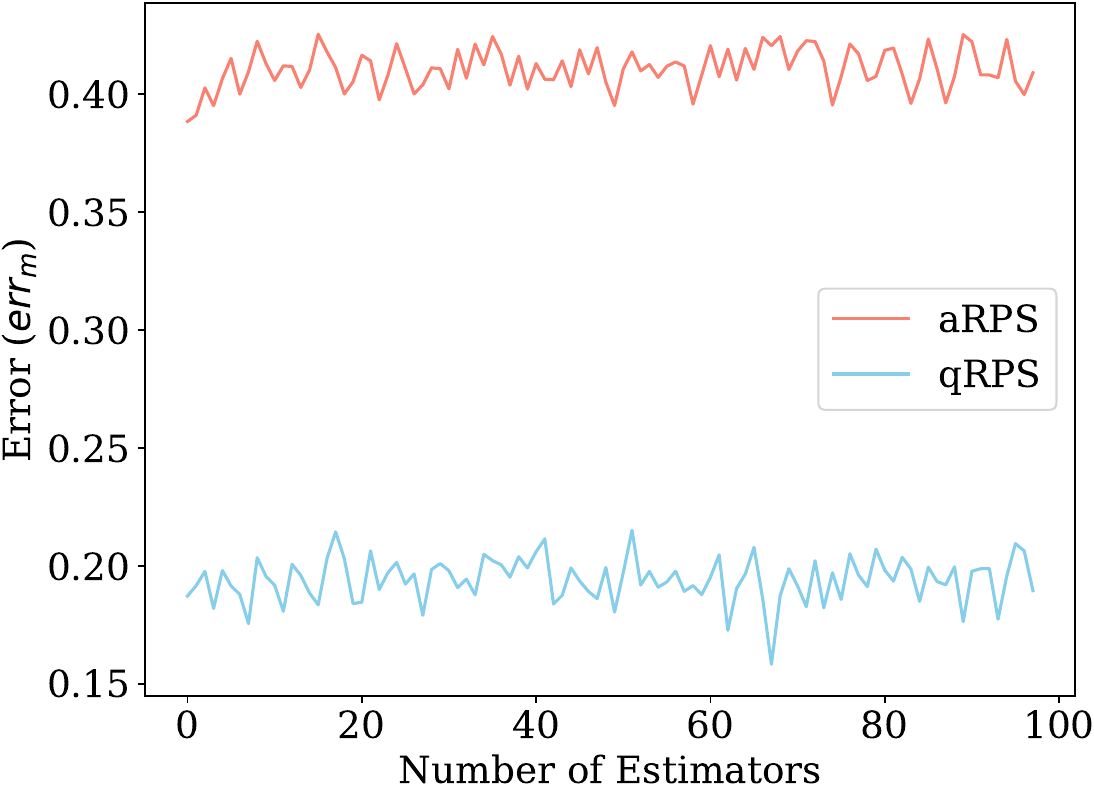}
  \caption{Error ($err_m$).}
  \label{fig:qrps_vs_arps_a}
\end{subfigure}\hfill
\begin{subfigure}{.49\linewidth}
  \includegraphics[width=\linewidth]{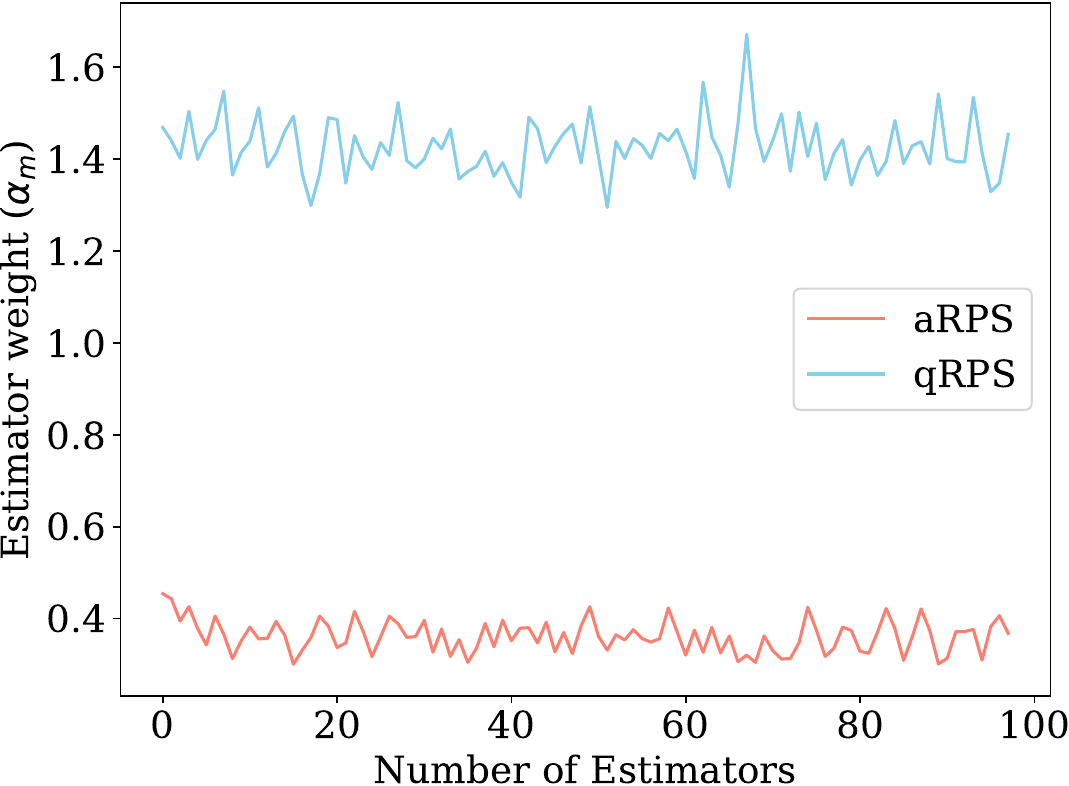}
  \caption{Estimator weight ($\alpha$).}
   \label{fig:qrps_vs_arps_b}
\end{subfigure}\\[0.5cm]
\begin{subfigure}{.49\linewidth}
  \includegraphics[width=\linewidth]{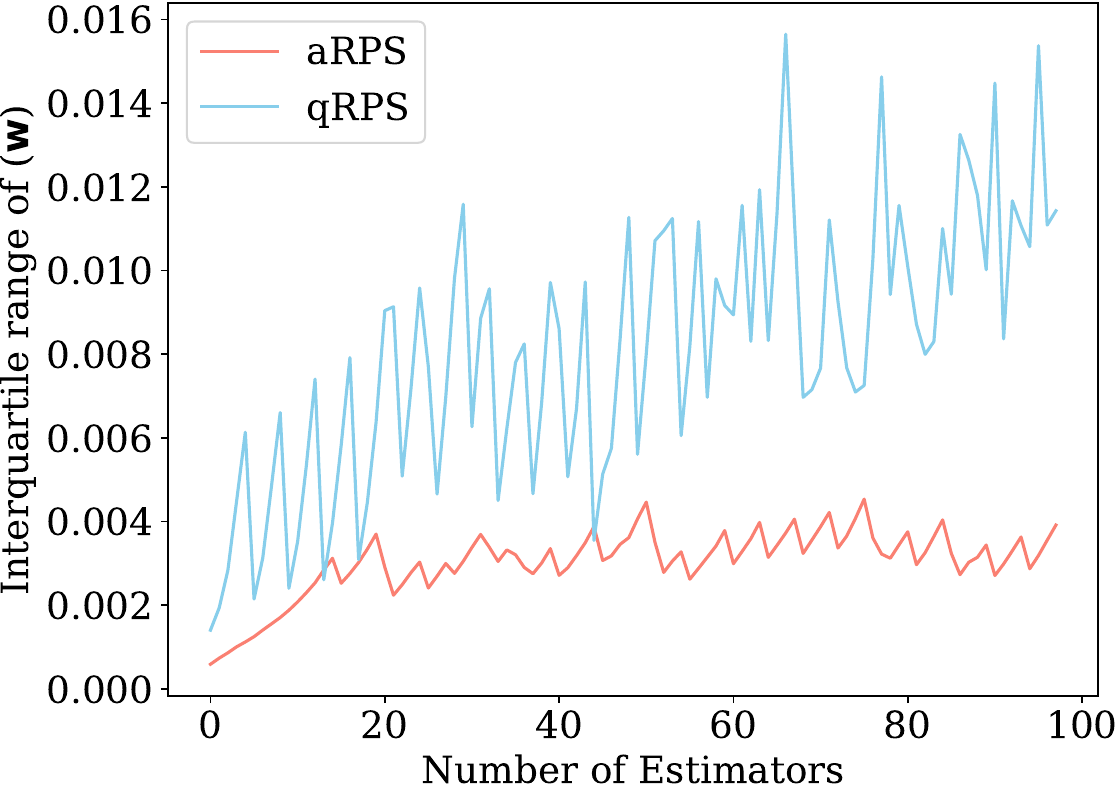}
  \caption{Interquartile range of sample weights ($\mathbf{w}$).}
   \label{fig:qrps_vs_arps_c}
\end{subfigure}
\begin{subfigure}{.49\linewidth}
  \includegraphics[width=\linewidth]{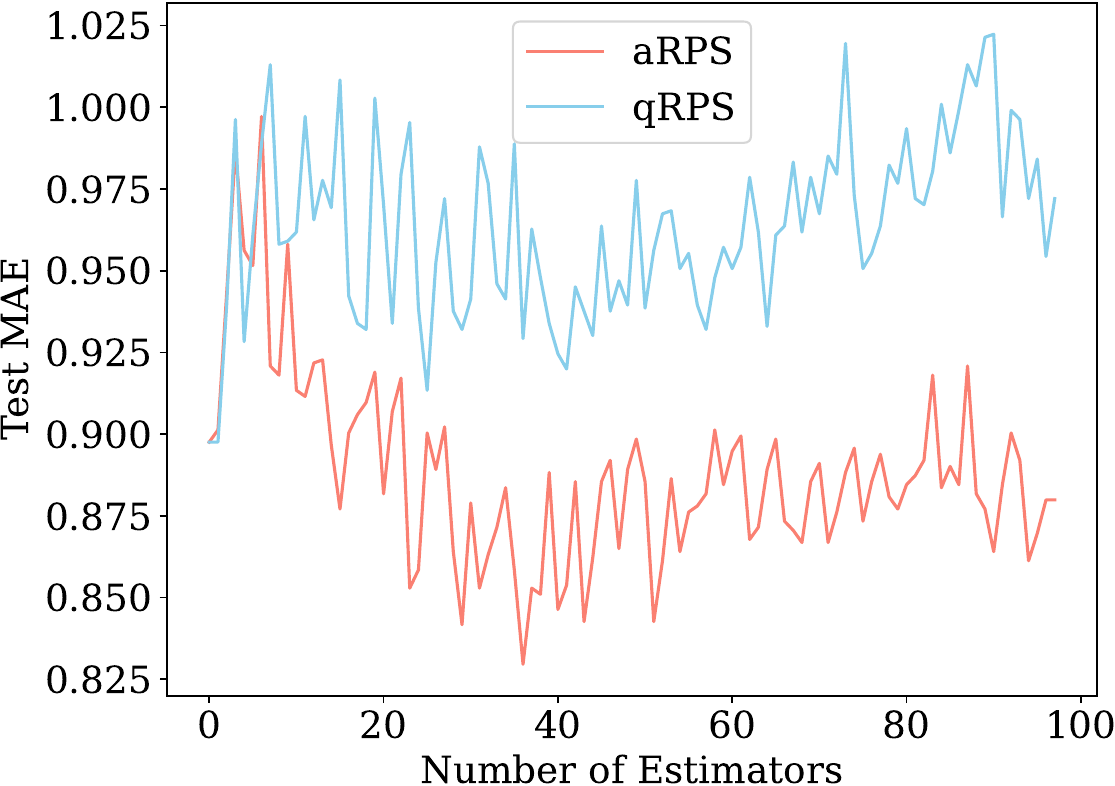}
  \caption{Mean Absolute Error (MAE) in the test set.}
  \label{fig:qrps_vs_arps_d}
\end{subfigure}
\caption{The absolute RPS (aRPS) and the quadratic RPS (qRPS) are compared by visualising the evolution of: (a) the estimator error $err_m$, (b) estimator weight $\alpha_m$, (c) interquartile range of the sample weights $\mathbf{w}$, and (d) MAE in the test set. The problem considered is \textit{oc04\_LESTSensors} (TOC-UCO repository \cite{tocuco}).}
\label{fig:qrps_vs_arps}
\end{figure}

A significant problem has been observed with the original qRPS, which motivates the use of the aRPS error function instead. This problem is illustrated in \Cref{fig:qrps_vs_arps}, where it can be observed that qRPS produces very low errors (\Cref{fig:qrps_vs_arps_a}), that lead to higher $\alpha$ values (\Cref{fig:qrps_vs_arps_b}), and ultimately induces a highly skewed sample weight distribution (\Cref{fig:qrps_vs_arps_c}), thus hindering the model convergence (\Cref{fig:qrps_vs_arps_d}). Two reasons are behind this fact. First, squaring amplifies the contribution of larger errors disproportionately, making the metric overly sensitive to outliers. Secondly, squared error does not respect the simplex structure of discrete probability distributions and may yield gradients that are ill-suited for learning algorithms operating under probability constraints. In contrast, the absolute difference treats deviations in a more balanced and uniform manner, all discrepancies contributing linearly to the total error, which makes the metric more robust to individual mismatches.

\subsection{Limitations and future work} \label{sec:limitations}

Although \texttt{ADABORD} provides significant contributions to OC, the present work has certain limitations. Despite the fact that the results demonstrated that the use of aRPS significantly enhances training robustness, which was also highlighted by Gneiting and Raftery in \cite{gneiting2007strictly}, the broader challenge of training robustness under extreme class imbalance remains relevant for further investigation \cite{bonnier22a}. A second limitation is related to the performance of the method across some classification metrics. While \texttt{ADABORD} demonstrated significant superiority in ordinal metrics, specifically Averaged Mean Absolute Error (AMAE) and Maximum Mean Absolute Error (MMAE), the comprehensive experimental study revealed that no single method universally outperforms the others across all metrics, which is in line with the results shown in recent literature reviews in ordinal classification such as \cite{tocuco,cardoso2025unimodal}. Notably, for non-ordinal metrics such as Balanced Accuracy (BACC), XGBoost achieved competitive or superior performance when compared to \texttt{ADABORD}. This is the main limitation of \texttt{ADABORD}, as both the base classifier and the boosting error function explicitly exploit the ordering of the target variable. Consequently, the application of the proposed approach to problems in which the target variable does not exhibit a pronounced ordinality is likely to result in inferior performance compared to nominal approaches \cite{lattke2015detecting,bellmann2020ordinal}.

Building upon the development of \texttt{ADABORD}, several promising future lines of research emerge. Firstly, to study the adaption to OC of other boosting approaches, such as XGBoost. Secondly, the application of these ordinal boosting techniques on a wider range of high-dimensional or time series ordinal datasets, particularly those from complex real-world application domains like medical diagnostics or financial risk assessment with $5$ or more classes. Thirdly, it is believed that adopting a class-averaged aRPS error function could further improve the performance of the proposed \texttt{ADABORD} approach on highly imbalanced datasets. Finally, novel approaches such as the DC-KNN algorithm \cite{sabri2024novel} could be adopted for ordinal classification. DC-KNN operates in two phases: a dynamic clustering step with adaptive distances and a $K$-nearest neighbour classifier trained on the augmented label space. Extending this framework to OC would require incorporating ordinal information into both phases, and the resulting classifier could be used to produce more informative and ordinally consistent base learners.

\section{Conclusions} \label{sec:conclusions}
In this work, \texttt{ADABORD}, a novel AdaBoost approach for Ordinal Classification (OC) problems, has been introduced. \texttt{ADABORD} extends the multiclass AdaBoost algorithm to effectively leverage the ordinal information. To this end, two key changes were introduced: 1) instead of the standard Decision Tree Classifier (DTC), an ordinal variant was introduced that uses the Ordinal Gini (OGini) splitting criterion, which has been proven to enhance performance on ordinal problems; and 2) the original error function of AdaBoost measuring the misclassification rate is replaced by an ordinal metric, which is also introduced in this work, named the absolute Ranked Probability Score (aRPS).

The rationale behind these two key modifications lies in the fact that the aRPS quantifies the divergence between the predicted and empirical cumulative distribution functions (cdfs), explicitly accounting for both the ordering and the relative distances between classes. The use of aRPS improves the robustness of the training process by mitigating issues observed with the quadratic RPS (qRPS), which can induce highly skewed sample-weight distributions and, consequently, hinder model convergence. In addition, the OGini splitting criterion exhibits a strong synergy with the aRPS, as it explicitly considers the cumulative class frequencies at each split. This alignment between the OGini criterion and the aRPS-based error function significantly enhances the convergence behaviour of the algorithm in ordinal classification problems.

A comprehensive experimental study was conducted to evaluate \texttt{ADABORD} against seven state-of-the-art approaches, including OEAdaBoost, AdaBoost, XGBoost, O-LGBM, BP-MLP, BP-MLP-CLM, and Ridge, across the largest OC benchmarking repository to date: the TOC-UCO repository, comprising $46$ ordinal datasets from diverse application domains. The results were statistically analysed, with superior performance by \texttt{ADABORD} being demonstrated on datasets with 5 or more classes, where ordinality is more pronounced.

\section*{Acknowledgments}
The present study has been supported by the ``Agencia Estatal de Investigación (España)'' (grant ref.: PID2023-150663NB-C22 / AEI / 10.13039 / 501100011033), by the European Commission, AgriFoodTEF (grant ref.: DI\-GI\-TAL-2022-CLOUD-AI-02, 101100622), by the Secretary of State for Digitalization and Artificial Intelligence ENIA International Chair (grant ref.: TSI-100921-2023-3), and by the University of Córdoba and Junta de Andalucía (grant ref.: PP2F\_L1\_15). R. Ayllón-Gavilán has been supported by the ``Instituto de Salud Carlos III'' (ISCIII) and EU (grant ref.: FI23/00163).

\bibliographystyle{elsarticle-num-names}
\bibliography{bibliography}

@article{gutierrez2016OrdinalRegressionMethods,
  title = {Ordinal {{Regression Methods}}: {{Survey}} and {{Experimental Study}}},
  author = {Guti{\'e}rrez, Pedro Antonio and {P{\'e}rez-Ortiz}, Mar{\'i}a and {S{\'a}nchez-Monedero}, Javier and {Fern{\'a}ndez-Navarro}, Francisco and {Herv{\'a}s-Mart{\'i}nez}, C{\'e}sar},
  year = {2016},
  journal = {IEEE Transactions on Knowledge and Data Engineering},
  volume = {28},
  number = {1},
  pages = {127--146},
  doi = {10.1109/TKDE.2015.2457911}
}

@article{TOLEDOCORTES2022105472,
title = {Grading diabetic retinopathy and prostate cancer diagnostic images with deep quantum ordinal regression},
journal = {Computers in Biology and Medicine},
volume = {145},
pages = {105472},
year = {2022},
issn = {0010-4825},
doi = {10.1016/j.compbiomed.2022.105472},
author = {Santiago Toledo-Cortés and Diego H. Useche and Henning Müller and Fabio A. González}
}

@article{GOLDMANN20241111,
title = {A new ordinal mixed-data sampling model with an application to corporate credit rating levels},
journal = {European Journal of Operational Research},
volume = {314},
number = {3},
pages = {1111-1126},
year = {2024},
issn = {0377-2217},
doi = {10.1016/j.ejor.2023.10.017},
author = {Leonie Goldmann and Jonathan Crook and Raffaella Calabrese}
}

@article{freund1997decision,
  title={A decision-theoretic generalization of on-line learning and an application to boosting},
  author={Freund, Yoav and Schapire, Robert E},
  journal={Journal of computer and system sciences},
  volume={55},
  number={1},
  pages={119--139},
  year={1997},
  publisher={Elsevier},
  doi={10.1006/jcss.1997.1504},
}

@article{zhu2009multi,
  title={Multi-class adaboost},
  author={Zhu, Ji and Zou, Hui and Rosset, Saharon and Hastie, Trevor and others},
  journal={Statistics and its Interface},
  volume={2},
  number={3},
  pages={349--360},
  year={2009},
  doi={10.4310/sii.2009.v2.n3.a8},
}

@article{um2023adaptive,
  title={Adaptive boosting for ordinal target variables using neural networks},
  author={Um, Insung and Lee, Geonseok and Lee, Kichun},
  journal={Statistical Analysis and Data Mining: The ASA Data Science Journal},
  volume={16},
  number={3},
  pages={257--271},
  year={2023},
  publisher={Wiley Online Library},
  doi={10.1002/sam.11613},
}

@book{hastie2009elements,
  title     = {The Elements of Statistical Learning: Data Mining, Inference, and Prediction},
  author    = {Hastie, Trevor and Tibshirani, Robert and Friedman, Jerome},
  year      = {2009},
  publisher = {Springer},
  series    = {Springer Series in Statistics},
  edition   = {2nd},
  address   = {New York},
  isbn      = {978-0-387-84857-0},
  doi       = {10.1007/978-0-387-21606-5}
}

@article{riccardi2014cost,
  title={Cost-sensitive AdaBoost algorithm for ordinal regression based on extreme learning machine},
  author={Riccardi, Annalisa and Fern{\'a}ndez-Navarro, Francisco and Carloni, Sante},
  journal={IEEE transactions on cybernetics},
  volume={44},
  number={10},
  pages={1898--1909},
  year={2014},
  publisher={IEEE},
  doi={10.1109/tcyb.2014.2299291},
}

@article{singer2020ordinal,
  title={Ordinal decision-tree-based ensemble approaches: The case of controlling the daily local growth rate of the {COVID-19} epidemic},
  author={Singer, Gonen and Marudi, Matan},
  journal={Entropy},
  volume={22},
  number={8},
  pages={871},
  year={2020},
  publisher={MDPI},
  doi={10.3390/e22080871},
}

@article{piccarreta2008classification,
  title={Classification trees for ordinal variables},
  author={Piccarreta, Raffaella},
  journal={Computational Statistics},
  volume={23},
  number={3},
  pages={407--427},
  year={2008},
  publisher={Springer},
  doi={10.1007/s00180-007-0077-5},
}

@article{ayllon2024splitting,
title = {Splitting criteria for ordinal decision trees: An experimental study},
journal = {Pattern Recognition},
volume = {171},
pages = {112273},
year = {2026},
issn = {0031-3203},
doi = {10.1016/j.patcog.2025.112273},
author = {Rafael Ayllón-Gavilán and Francisco José Martínez-Estudillo and David Guijo-Rubio and César Hervás-Martínez and Pedro Antonio Gutiérrez},
}

@article{tocuco,
	author = {Ayll{\' o}n-Gavil{\' a}n, Rafael and Guijo-Rubio, David and G{\' o}mez-Orellana, Antonio Manuel and B{\' e}rchez-Moreno, Francisco and Vargas-Yun, V{\' i}ctor Manuel and Guti{\' e}rrez, Pedro Antonio},
    journal = {Neurocomputing},
	doi = {10.1016/j.neucom.2026.133528},
	year = {2026},
	pages = {133528},
	title = {{TOC}-{UCO}: a comprehensive repository of tabular ordinal classification datasets},
}

@inproceedings{xgboost,
author = {Chen, Tianqi and Guestrin, Carlos},
title = {XGBoost: A Scalable Tree Boosting System},
year = {2016},
isbn = {9781450342322},
publisher = {Association for Computing Machinery},
address = {New York, NY, USA},
doi = {10.1145/2939672.2939785},
booktitle = {Proceedings of the 22nd ACM SIGKDD International Conference on Knowledge Discovery and Data Mining},
pages = {785–794},
numpages = {10},
location = {San Francisco, California, USA},
series = {KDD '16}
}

@article{CRUZRAMIREZ201421,
title = {Metrics to guide a multi-objective evolutionary algorithm for ordinal classification},
journal = {Neurocomputing},
volume = {135},
pages = {21-31},
year = {2014},
issn = {0925-2312},
doi = {10.1016/j.neucom.2013.05.058},
author = {M. Cruz-Ramírez and C. Hervás-Martínez and J. Sánchez-Monedero and P.A. Gutiérrez},
}

@article{middlehurst2024aeon,
  title={aeon: a Python toolkit for learning from time series},
  author={Middlehurst, Matthew and Ismail-Fawaz, Ali and Guillaume, Antoine and Holder, Christopher and Guijo-Rubio, David and Bulatova, Guzal and Tsaprounis, Leonidas and Mentel, Lukasz and Walter, Martin and Sch{\"a}fer, Patrick and others},
  journal={Journal of Machine Learning Research},
  volume={25},
  number={289},
  pages={1--10},
  year={2024}
}

@article{epstein1969scoring,
  title={A scoring system for probability forecasts of ranked categories},
  author={Epstein, Edward S},
  journal={Journal of Applied Meteorology (1962-1982)},
  volume={8},
  number={6},
  pages={985--987},
  year={1969},
  publisher={JSTOR},
  doi={10.1175/1520-0450(1969)008<0985:assfpf>2.0.co;2},
}

@article{guijo2018prediction,
  title={Prediction of low-visibility events due to fog using ordinal classification},
  author={Guijo-Rubio, David and Guti{\'e}rrez, PA and Casanova-Mateo, Carlos and Sanz-Justo, Julia and Salcedo-Sanz, Sancho and Herv{\'a}s-Mart{\'\i}nez, C{\'e}sar},
  journal={Atmospheric Research},
  volume={214},
  pages={64--73},
  year={2018},
  doi={10.1016/j.atmosres.2018.07.017},
  publisher={Elsevier}
}

@article{gomez2024orfeo,
  title={{ORFEO}: Ordinal classifier and regressor fusion for estimating an ordinal categorical target},
  author={G{\'o}mez-Orellana, Antonio M and Guijo-Rubio, David and Guti{\'e}rrez, Pedro A and Herv{\'a}s-Mart{\'\i}nez, C{\'e}sar and Vargas, V{\'\i}ctor M},
  journal={Engineering Applications of Artificial Intelligence},
  volume={133},
  pages={108462},
  year={2024},
  publisher={Elsevier},
  doi={10.1016/j.engappai.2024.108462},
}

@article{duran2021ordinal,
  title={Ordinal classification of the affectation level of 3D-images in Parkinson diseases},
  author={Dur{\'a}n-Rosal, Antonio M and Camacho-Ca{\~n}am{\'o}n, Julio and Guti{\'e}rrez, Pedro Antonio and Guiote Moreno, Maria Victoria and Rodr{\'\i}guez-C{\'a}ceres, Ester and Vallejo Casas, Juan Antonio and Herv{\'a}s-Mart{\'\i}nez, C{\'e}sar},
  journal={Scientific Reports},
  volume={11},
  number={1},
  pages={7067},
  year={2021},
  doi={10.1038/s41598-021-86538-y},
  publisher={Nature Publishing Group UK London}
}

@article{solares2022multicriteria,
  title={Multicriteria ordinal classification to improve strategic planning in the financial sector of the company},
  author={Solares, Efra{\'\i}n and de-Le{\'o}n-G{\'o}mez, V{\'\i}ctor and Fern{\'a}ndez, Eduardo and Contreras-Medina, Emmanuel and Lopez, Orlando},
  journal={International Journal of Combinatorial Optimization Problems and Informatics},
  volume={13},
  number={2},
  pages={47},
  year={2022},
  publisher={International Journal of Combinatorial Optimization Problems \& Informatics},
  doi={10.61467/2007.1558.2022.v13i2.270},
}

@article{lazaro2023neural,
  title={Neural network for ordinal classification of imbalanced data by minimizing a Bayesian cost},
  author={L{\'a}zaro, Marcelino and Figueiras-Vidal, An{\'\i}bal R},
  journal={Pattern Recognition},
  volume={137},
  pages={109303},
  year={2023},
  publisher={Elsevier},
  doi={10.1016/j.patcog.2023.109303},
}

@inproceedings{baccianella2009evaluation,
  title={Evaluation measures for ordinal regression},
  author={Baccianella, Stefano and Esuli, Andrea and Sebastiani, Fabrizio},
  booktitle={2009 Ninth international conference on intelligent systems design and applications},
  pages={283--287},
  year={2009},
  organization={IEEE},
  doi={10.1109/isda.2009.230},
}

@book{agresti2010analysis,
  title={Analysis of ordinal categorical data},
  author={Agresti, Alan},
  year={2010},
  publisher={John Wiley \& Sons}
}

@article{perez2013projection,
  title={Projection-based ensemble learning for ordinal regression},
  author={P{\'e}rez-Ort{\'i}z, Mar{\'i}a and Guti{\'e}rrez, Pedro Antonio and Herv{\'a}s-Mart{\'\i}nez, C{\'e}sar},
  journal={IEEE transactions on cybernetics},
  volume={44},
  number={5},
  pages={681--694},
  year={2013},
  publisher={IEEE},
  doi={10.1109/tcyb.2013.2266336},
}

@article{fernandez2013negative,
  title={Negative correlation ensemble learning for ordinal regression},
  author={Fern{\'a}ndez-Navarro, Francisco and Guti{\'e}rrez, Pedro Antonio and Herv{\'a}s-Mart{\'a}nez, C{\'e}sar and Yao, Xin},
  journal={IEEE transactions on neural networks and learning systems},
  volume={24},
  number={11},
  pages={1836--1849},
  year={2013},
  publisher={IEEE},
  doi={10.1109/tnnls.2013.2268279},
}

@article{guo2021ensemble,
  title={An ensemble learning method based on ordinal regression for {COVID-19} diagnosis from chest {CT}},
  author={Guo, Xiaodong and Lei, Yiming and He, Peng and Zeng, Wenbing and Yang, Ran and Ma, Yinjin and Feng, Peng and Lyu, Qing and Wang, Ge and Shan, Hongming},
  journal={Physics in Medicine \& Biology},
  volume={66},
  number={24},
  pages={244001},
  year={2021},
  publisher={IOP Publishing},
  doi={10.1088/1361-6560/ac34b2 },
}

@article{okuno2024interpretable,
  title={An Interpretable Neural Network-based Nonproportional Odds Model for Ordinal Regression},
  author={Okuno, Akifumi and Harada, Kazuharu},
  journal={Journal of Computational and Graphical Statistics},
  volume={33},
  number={4},
  pages={1454--1463},
  year={2024},
  publisher={Taylor \& Francis},
  doi={10.1080/10618600.2024.2321208},
}

@article{marudi2024decision,
  title={A decision tree-based method for ordinal classification problems},
  author={Marudi, Matan and Ben-Gal, Irad and Singer, Gonen},
  journal={IISE Transactions},
  volume={56},
  number={9},
  pages={960--974},
  year={2024},
  publisher={Taylor \& Francis},
  doi={10.1080/24725854.2022.2081745},
}

@article{ghasemkhani2025ordinal,
  title={Ordinal Random Tree with Rank-Oriented Feature Selection ({ORT-ROFS}): A Novel Approach for the Prediction of Road Traffic Accident Severity},
  author={Ghasemkhani, Bita and Balbal, Kadriye Filiz and Kokten, Ulas Birant and Birant, Derya},
  journal={Mathematics},
  volume={13},
  number={2},
  pages={310},
  year={2025},
  publisher={MDPI AG},
  doi={10.3390/math13020310}
}

@article{vega2021ocean,
  title={{OCEAn}: Ordinal classification with an ensemble approach},
  author={Vega-M{\'a}rquez, Bel{\'e}n and Nepomuceno-Chamorro, Isabel A and Rubio-Escudero, Cristina and Riquelme, Jos{\'e} C},
  journal={Information Sciences},
  volume={580},
  pages={221--242},
  year={2021},
  publisher={Elsevier},
  doi={10.1016/j.ins.2021.08.081},
}

@inproceedings{galdran2023performance,
  title={Performance metrics for probabilistic ordinal classifiers},
  author={Galdrán, Adrian},
  booktitle={International Conference on Medical Image Computing and Computer-Assisted Intervention},
  pages={357--366},
  year={2023},
  organization={Springer},
  doi={10.1007/978-3-031-43898-1_35},
}

@article{vargas2024ebano,
  title={{EBANO}: A novel Ensemble BAsed on uNimodal Ordinal classifiers for the prediction of significant wave height},
  author={Vargas, V{\'\i}ctor M and G{\'o}mez-Orellana, Antonio M and Guti{\'e}rrez, Pedro A and Herv{\'a}s-Mart{\'\i}nez, C{\'e}sar and Guijo-Rubio, David},
  journal={Knowledge-Based Systems},
  volume={300},
  pages={112223},
  year={2024},
  publisher={Elsevier},
  doi={10.1016/j.knosys.2024.112223},
}

@article{delgado2025ordmap,
title = {Ord-MAP criterion: Extending MAP for ordinal classification},
journal = {Knowledge-Based Systems},
volume = {324},
pages = {113837},
year = {2025},
issn = {0950-7051},
doi = {10.1016/j.knosys.2025.113837},
author = {Rosario Delgado}
}

@article{pelaez2024general,
  title={A general explicable forecasting framework for weather events based on ordinal classification and inductive rules combined with fuzzy logic},
  author={Pel{\'a}ez-Rodr{\'\i}guez, C{\'e}sar and P{\'e}rez-Aracil, Jorge and Marina, Cosmin Madalin and Prieto-Godino, Luis and Casanova-Mateo, Carlos and Guti{\'e}rrez, Pedro Antonio and Salcedo-Sanz, Sancho},
  journal={Knowledge-Based Systems},
  volume={291},
  pages={111556},
  year={2024},
  publisher={Elsevier},
  doi={10.1016/j.knosys.2024.111556},
}

@article{kucuksari2023new,
  title={A new rough ordinal priority-based decision support system for purchasing electric vehicles},
  author={Kucuksari, Sadik and Pamucar, Dragan and Deveci, Muhammet and Erdogan, Nuh and Delen, Dursun},
  journal={Information Sciences},
  volume={647},
  pages={119443},
  year={2023},
  publisher={Elsevier}, 
  doi={10.1016/j.ins.2023.119443}
}

@inproceedings{perez2016ordinalimbalance,
  title = {Tackling the Ordinal and Imbalance Nature of a Melanoma Image Classification Problem},
  booktitle = {2016 International Joint Conference on Neural Networks ({{IJCNN}})},
  author = {{P{\'e}rez-Ortiz}, M. and S{\'a}ez, A. and {S{\'a}nchez-Monedero}, J. and Guti{\'e}rrez, P.A. and {Herv{\'a}s-Mart{\'i}nez}, C.},
  year = {2016},
  pages = {2156--2163},
  doi = {10.1109/IJCNN.2016.7727466}
}

@article{hu2008adaboost,
  title={Adaboost-based algorithm for network intrusion detection},
  author={Hu, Weiming and Hu, Wei and Maybank, Steve},
  journal={IEEE Transactions on Systems, Man, and Cybernetics, Part B (Cybernetics)},
  volume={38},
  number={2},
  pages={577--583},
  year={2008},
  publisher={IEEE},
  doi={10.1109/TSMCB.2007.914695}
}

@article{cohen1968weighted,
  title={Weighted kappa: Nominal scale agreement provision for scaled disagreement or partial credit.},
  author={Cohen, Jacob},
  journal={Psychological bulletin},
  volume={70},
  number={4},
  pages={213},
  year={1968},
  publisher={American Psychological Association},
  doi={10.1037/h0026256},
}

@article{warrens2012cohen,
  title={Cohen’s quadratically weighted kappa is higher than linearly weighted kappa for tridiagonal agreement tables},
  author={Warrens, Matthijs J},
  journal={Statistical Methodology},
  volume={9},
  number={3},
  pages={440--444},
  year={2012},
  publisher={Elsevier},
  doi={10.1016/j.stamet.2011.08.006}
}

@inproceedings{brodersen2010balanced,
  title={The balanced accuracy and its posterior distribution},
  author={Brodersen, Kay Henning and Ong, Cheng Soon and Stephan, Klaas Enno and Buhmann, Joachim M},
  booktitle={2010 20th international conference on pattern recognition},
  pages={3121--3124},
  year={2010},
  organization={IEEE},
  doi={10.1109/ICPR.2010.764}
}

@book{bishop2006mlp,
  address =       {Secaucus, NJ, USA},
  author =        {Bishop, C. M.},
  publisher =     {Springer-Verlag New York, Inc.},
  title =         {Pattern Recognition and Machine Learning (Information Science and Statistics)},
  year =          {2006},
  isbn =          {0387310738},
}

@article{pedregosa2011scikit,
  title={Scikit-learn: Machine learning in Python},
  author={Pedregosa, Fabian and Varoquaux, Ga{\"e}l and Gramfort, Alexandre and Michel, Vincent and Thirion, Bertrand and Grisel, Olivier and Blondel, Mathieu and Prettenhofer, Peter and Weiss, Ron and Dubourg, Vincent and others},
  journal={the Journal of machine Learning research},
  volume={12},
  pages={2825--2830},
  year={2011},
  publisher={JMLR. org}
}

@article{breiman2001random,
  title={Random forests},
  author={Breiman, Leo},
  journal={Machine learning},
  volume={45},
  number={1},
  pages={5--32},
  year={2001},
  publisher={Springer},
  doi={10.1023/a:1010933404324},
}

@inproceedings{cheng2008neural,
  title={A neural network approach to ordinal regression},
  author={Cheng, Jianlin and Wang, Zheng and Pollastri, Gianluca},
  booktitle={2008 IEEE international joint conference on neural networks (IEEE world congress on computational intelligence)},
  pages={1279--1284},
  year={2008},
  organization={IEEE},
  doi={10.1109/IJCNN.2008.4633963},
}

@article{gneiting2007strictly,
  title={Strictly proper scoring rules, prediction, and estimation},
  author={Gneiting, Tilmann and Raftery, Adrian E},
  journal={Journal of the American statistical Association},
  volume={102},
  number={477},
  pages={359--378},
  year={2007},
  publisher={Taylor \& Francis},
  doi={10.1198/016214506000001437},
}

@article{shahri2022novel,
  title={A novel approach to uncertainty quantification in groundwater table modeling by automated predictive deep learning},
  author={Shahri, Abbas Abbaszadeh and Shan, Chunling and Larsson, Stefan},
  journal={Natural Resources Research},
  volume={31},
  number={3},
  pages={1351--1373},
  year={2022},
  publisher={Springer Nature BV},
  doi={10.1007/s11053-022-10051-w},
}

@article{altalhan2025imbalanced,
  title={Imbalanced data problem in machine learning: A review},
  author={Altalhan, Manahel and Algarni, Abdulmohsen and Alouane, Monia Turki-Hadj},
  journal={IEEE Access},
  year={2025},
  publisher={IEEE},
  doi={10.1109/ACCESS.2025.3531662}
}

@article{delatorreQWK2018,
title = {Weighted kappa loss function for multi-class classification of ordinal data in deep learning},
journal = {Pattern Recognition Letters},
volume = {105},
pages = {144-154},
year = {2018},
issn = {0167-8655},
doi = {https://doi.org/10.1016/j.patrec.2017.05.018},
author = {Jordi {de la Torre} and Domenec Puig and Aida Valls},
}

@inproceedings{lattke2015detecting,
  title={Detecting ordinal class structures},
  author={Lattke, Raphael and Lausser, Ludwig and M{\"u}ssel, Christoph and Kestler, Hans A},
  booktitle={International Workshop on Multiple Classifier Systems},
  pages={100--111},
  year={2015},
  organization={Springer},
  doi={10.1007/978-3-319-20248-8_9}
}

@article{bellmann2020ordinal,
  title={Ordinal classification: Working definition and detection of ordinal structures},
  author={Bellmann, Peter and Schwenker, Friedhelm},
  journal={IEEE Access},
  volume={8},
  pages={164380--164391},
  year={2020},
  publisher={IEEE},
  doi={10.1109/ACCESS.2020.3021596}
}

@InProceedings{bonnier22a,
  title = 	 {Assessing the Robustness of Ordinal Classifiers against Imbalanced and Shifting Distributions},
  author =       {Bonnier, Thomas and Bosch, Benjamin},
  booktitle = 	 {Proceedings of the Fourth International Workshop on Learning with Imbalanced Domains: Theory and Applications},
  pages = 	 {112--126},
  year = 	 {2022},
  editor = 	 {Moniz, Nuno and Branco, Paula and Torgo, Luís and Japkowicz, Nathalie and Wozniak, Michal and Wang, Shuo},
  volume = 	 {183},
  series = 	 {Proceedings of Machine Learning Research},
  month = 	 {23 Sep},
  publisher =    {PMLR},
  url = 	 {https://proceedings.mlr.press/v183/bonnier22a.html},
}

@article{rumelhart1986learning,
  title={Learning representations by back-propagating errors},
  author={Rumelhart, David E and Hinton, Geoffrey E and Williams, Ronald J},
  journal={nature},
  volume={323},
  number={6088},
  pages={533--536},
  year={1986},
  publisher={Nature Publishing Group UK London},
  doi={10.1038/323533a0}
}

@inproceedings{diederik2015adam,
      title={Adam: A Method for Stochastic Optimization}, 
      author={Diederik P. Kingma, Jimmy Ba},
      booktitle={International Conference for Learning Representations},
      year={2015},
      pages={1--15},
}

@article{cardoso2025unimodal,
  title={Unimodal distributions for ordinal regression},
  author={Cardoso, Jaime S and Cruz, Ricardo PM and Albuquerque, Tom{\'e}},
  journal={IEEE Transactions on Artificial Intelligence},
  year={2025},
  publisher={IEEE},
  doi={10.1109/TAI.2025.3549740},
}

@article{mirfallah2025geothermal,
  title={Geothermal resource classification in Catalonia (Spain) using AI-derived predictions},
  author={Mirfallah Lialestani, Seyed Poorya and Parcerisa, David and Himi, Mahjoub and Abbaszadeh Shahri, Abbas},
  journal={Energies},
  volume={18},
  number={22},
  pages={6040},
  year={2025},
  publisher={MDPI},
  doi={10.3390/en18226040}
}

@article{sabri2024novel,
  title={A novel classification algorithm based on the synergy between dynamic clustering with adaptive distances and k-nearest neighbors},
  author={Sabri, Mohammed and Verde, Rosanna and Balzanella, Antonio and Maturo, Fabrizio and Tairi, Hamid and Yahyaouy, Ali and Riffi, Jamal},
  journal={Journal of Classification},
  volume={41},
  number={2},
  pages={264--288},
  year={2024},
  publisher={Springer},
  doi={10.1007/s00357-024-09471-5}
}

@misc{spanashis2024ordinalgbt,
  author       = {Spanashis, Adamos},
  title        = {OrdinalGBT: A Library for Ordinal Gradient Boosted Trees},
  year         = {2024},
  url          = {https://ordinalgbt.readthedocs.io/en/latest/},
  note         = {Accessed: 2026-05-04}
}

@article{ke2017lightgbm,
  title={Lightgbm: A highly efficient gradient boosting decision tree},
  author={Ke, Guolin and Meng, Qi and Finley, Thomas and Wang, Taifeng and Chen, Wei and Ma, Weidong and Ye, Qiwei and Liu, Tie-Yan},
  journal={Advances in neural information processing systems},
  volume={30},
  year={2017},
  url={https://dl.acm.org/doi/10.5555/3294996.3295074}
}

@article{clm,
Author = {Vargas, Victor Manuel and Gutierrez, Pedro Antonio and Hervas-Martinez,
   Cesar},
Title = {Cumulative link models for deep ordinal classification},
Journal = {Neurocomputing},
Year = {2020},
Volume = {401},
Pages = {48-58},
DOI = {10.1016/j.neucom.2020.03.034},
ISSN = {0925-2312},
EISSN = {1872-8286},
ResearcherID-Numbers = {Gutiérrez, Pedro Antonio/K-6051-2014
   Hervas-Martinez, Cesar/A-3979-2009},
ORCID-Numbers = {Gutiérrez, Pedro Antonio/0000-0002-2657-776X
   Vargas Yun, Victor Manuel/0000-0002-0700-275X
   Hervas-Martinez, Cesar/0000-0003-4564-1816},
Unique-ID = {WOS:000544725700005},
}

\end{document}